\documentclass[10pt,twocolumn,letterpaper]{article}

\usepackage{cvpr}              % To produce the CAMERA-READY version
\definecolor{cvprblue}{rgb}{0.21,0.49,0.74}
\usepackage[pagebackref,breaklinks,colorlinks,allcolors=cvprblue]{hyperref}
\usepackage{multirow}
\usepackage{algorithm}
\usepackage{algpseudocode}
\usepackage{multirow}
\usepackage{makecell}
\usepackage{tabularx}

\def\paperID{15235} % *** Enter the Paper ID here
\def\confName{CVPR}
\def\confYear{2026}

\title{HiViS: Hiding Visual Tokens from the Drafter for Speculative Decoding in Vision-Language Models}

\author{Zhinan Xie\textsuperscript{1,2} \quad
Peisong Wang\textsuperscript{1,2,3} \quad
Shuang Qiu\textsuperscript{4} \quad
Jian Cheng\textsuperscript{1,2,3} \\
\textsuperscript{1}C$^2$DL, Institute of Automation, Chinese Academy of Sciences \\
\textsuperscript{2}School of Artificial Intelligence, University of Chinese Academy of Sciences \\
\textsuperscript{3}AIRIA \quad
\textsuperscript{4}City University of Hong Kong\\
}

\newcommand{\blfootnote}[1]{%
  \begingroup
  \renewcommand{\thefootnote}{}%
  \footnote{#1}%
  \addtocounter{footnote}{-1}%
  \endgroup
}

\begin{document}
\maketitle
\blfootnote{\textit{39th IEEE/CVF
Conference on Computer Vision and Pattern Recognition findings.}}
\begin{abstract}

Speculative decoding has proven effective for accelerating inference in Large Language Models (LLMs), yet its extension to Vision-Language Models (VLMs) remains limited by the computational burden and semantic inconsistency introduced by visual tokens. Recent studies reveal that visual tokens in large VLMs are highly redundant, and most of them can be removed without compromising generation quality. Motivated by this observation, we propose HiViS (\textbf{H}iding \textbf{V}isual Tokens from the Drafter for \textbf{S}peculative Decoding in Vision-Language Models), a framework that utilizes the target VLM as a semantic fusion model, allowing the drafter to obtain visual information without explicitly processing visual tokens, ensuring that the drafter's prefill sequence length matches that of the textual tokens. Furthermore, HiViS employs a time-step-aware aligned training scheme that allows the drafter to autonomously propagate and refine instructive visual-textual semantics during independent drafting, guided by step-dependent bias-correction residuals. Extensive experiments across representative VLMs and benchmarks demonstrate that HiViS achieves significant improvements in average acceptance length and speedup ratio. Our code is available at \url{https://github.com/lnn-ops/HiViS}.
% while maintaining high semantic consistency with the target VLM.
\end{abstract}    
\section{Introduction}
\label{sec:intro}
% Autoregressive (AR) models have emerged as the backbone of modern generative AI. Large language models (LLMs) \cite{achiam2023gpt,touvron2023llama} exemplify this trend, demonstrating strong language understanding and generation capabilities across diverse Natural Language Processing (NLP) tasks, while Vision-Language models (VLMs) \cite{li2023blip, liu2023llava,liu2024llavanext} extend the same AR principle to multimodal settings, enabling tasks such as image captioning and visual question answering. Despite their success, AR inference is inherently sequential: each token depends on all previously generated tokens. This property makes inference latency heavily constrained by memory bandwidth and communication, posing a significant bottleneck for both LLMs and VLMs.
Autoregressive (AR) models form the backbone of modern generative AI. Large Language Models (LLMs) \cite{achiam2023gpt, touvron2023llama} provide strong language understanding and generation capabilities, and Vision–Language Models (VLMs) \cite{li2023blip, liu2023llava, liu2024llavanext} extend the same AR formulation to multimodal tasks such as image understanding and visual question answering. Despite their effectiveness, AR inference is inherently sequential because each token depends on all previously generated ones, causing latency to be constrained by memory bandwidth and communication in both LLMs and VLMs.

\begin{figure}
    \centering
    \includegraphics[width=0.9\linewidth]{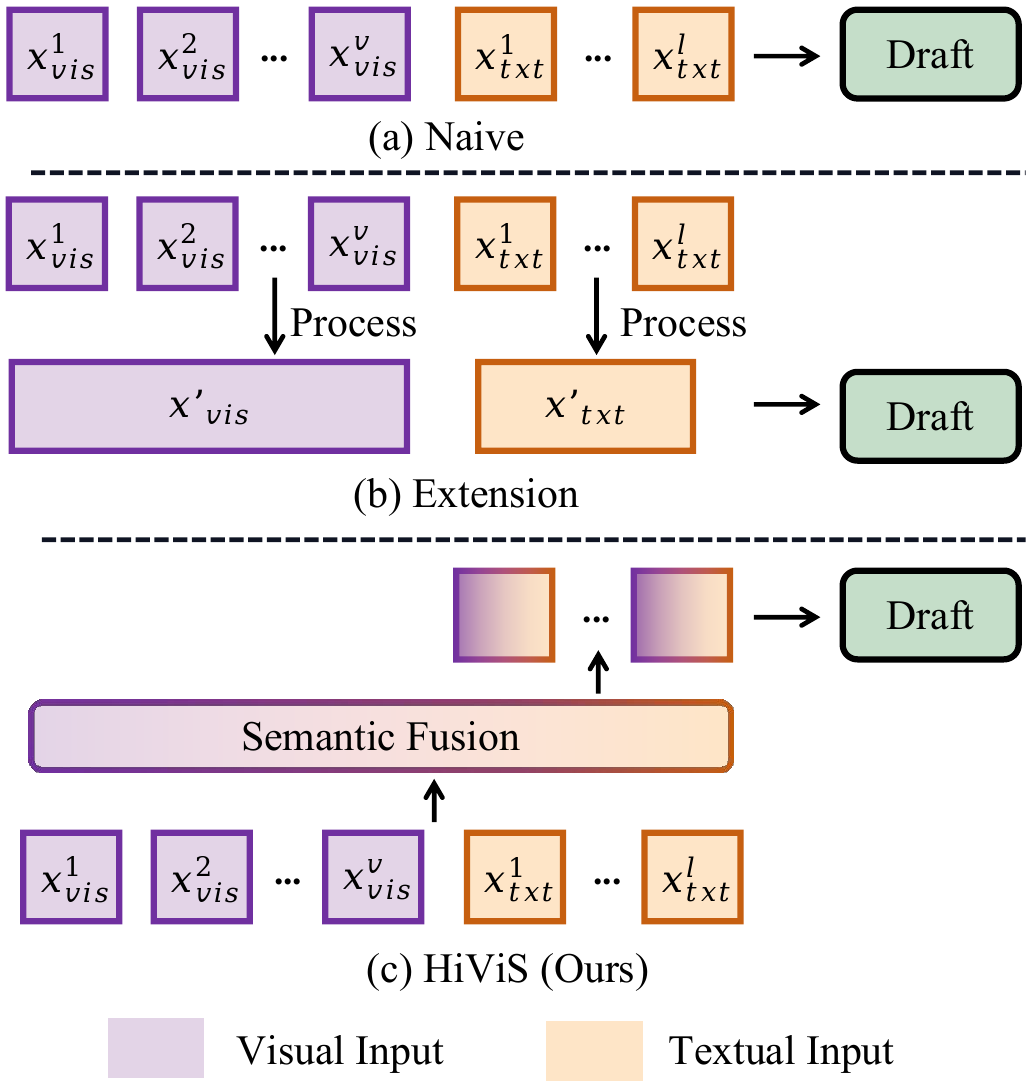}
    \caption{
    % Comparison of speculative decoding input modality for VLM drafters. The naive method directly reuses both visual and textual tokens from the target VLM. Extensions either process visual and textual tokens differently or remove visual tokens without other visual information. HiViS hides all visual tokens, retaining only text tokens as explicit input and augmenting them with implicit visual semantics.
    Input modalities for drafting.(a) Naive: feed both visual and textual tokens directly to the drafter. (b) Extensions: Additional module pre-processing visual and textual tokens before drafting. (c) HiViS: removes visual tokens and uses a semantic fusion module to enrich textual tokens with visual information.}
    \label{fig:var_input}
\end{figure}

Speculative decoding \cite{xia2022speculative,leviathan2023fast,chen2023accelerating} addresses the sequential inefficiency of autoregressive inference through a two-stage draft-verify process. A lightweight drafter first autoregressively generates candidate tokens, which the target model verifies in parallel within a single forward pass. Accepted tokens are retained, while rejected ones are resampled. This approach effectively reduces decoding steps while maintaining output quality.
%while preserving the exact output distribution of the target model. 
Although extensively explored in LLMs \cite{fu2024break, du2024glide,li2024eagle,sun2024triforce,cai2024medusa}, its application to Vision-Language Models (VLMs) remains under-explored. This gap is non-trivial: modern VLMs inherit the large-scale architecture of LLMs and thus suffer from the same inference inefficiency, while also introducing additional complexity and fundamental obstacles due to the presence of visual tokens:

% Forcing  drafter能力弱，
% co-adaptation

\textbf{The capacity and semantic gap between drafter and target model}.
The visual tokens produced by the target VLM's encoder are highly tailored to its own architecture and thus well aligned with target LLM's semantic representation. This cross-modal alignment is achieved by the co-adaptation between the high capacity target LLM and its visual encoder. Although carrying rich multimodal semantics, these visual tokens are not naturally aligned with the representation space of a lightweight drafter. Feeding them directly into the drafter distorts the contextual signals, because the drafter has no capacity to deal with these cross-modal feature fusion. As a result, the mismatch between the drafter and target model leads to increased distributional discrepancy and lowers the acceptance rate.  

\textbf{Computational burden of long visual token sequences.} 
Although the drafter remains lightweight to ensure efficiency, the requirement to handle long and complex visual token sequences increases its computational overhead. For instance, LLaVA-Next \cite{liu2024llavanext} produces over 2,000 visual tokens when processing high-resolution images, making visual inputs the dominant factor in sequence length and severely slowing down the drafter.

In this work, we propose \textbf{H}iding \textbf{V}isual Tokens from the Drafter for \textbf{S}peculative Decoding in Vision-Language Models (\textbf{HiViS}), a speculative decoding framework for VLMs that removes all visual tokens from the drafter yet still remaining visual information, as shown in \Cref{fig:var_input}(c). 
% We propose a hybrid multimodal semantic input mechanism, in which text tokens explicitly provide basic linguistic representations. Owing to the cross-modality self-attention mechanism, the target VLM naturally serves as a semantic fusion model, providing both visual and supplementary textual semantics. These are incorporated with text tokens via weighted addition as a fused embedding for the drafter. This mechanism effectively maintains multimodal consistency without necessitating explicit visual token processing by the drafter.
We leverage the target VLM as a natural semantic fusion model for its cross-modality self-attention. This allows us to obtain visual injected text embeddings containing both visual and textual semantics, which are utilized as the drafter's inputs, enabling the drafter to access multimodal information.
While this scheme constructs a compact and semantically consistent KV-cache that accelerates decoding, the drafter cannot directly access updated visual semantics from the target VLM during independent drafting. To address this, 
% the drafter is trained to update visual and textual information. A step-dependent bias correction residual is introduced to refine the propagated signals and alleviate accumulated errors. During training, we simulate the inference process to optimize residuals and drafter, and dynamically retain within-capacity yet challenging samples, improving training stability and efficiency. Consequently, the drafter can autonomously propagate and update visual and textual semantics.
we introduce a dynamic time-step aligned training mechanism in which a step-dependent bias correction residual is applied at each decoding step. This unified process adapts the visual injected text embeddings to the drafter and enables the drafter to propagate and update visual-textual semantics over multiple steps, enabling the drafter to operate autonomously without feedback from the target VLM.

In conclusion, our contributions are as follows:

\begin{itemize}
  \item We introduce a new paradigm for VLM speculative decoding in which the drafter avoids directly processing visual tokens, with support from the target VLM.
  \item We propose a training scheme that enables the drafter to implicitly propagate visual and textual information during independent drafting, augmented with step-dependent bias-correction residuals to support inference.
  \item Extensive experiments on representative VLMs and benchmarks demonstrate consistent gains in acceptance length and speedup (up to 3.15$\times$), while losslessly preserving the target VLM's output distribution.
\end{itemize}
\section{Related Works}
\label{sec:Preliminaries}

\subsection{Speculative Decoding}
% Speculative decoding accelerates autoregressive inference by employing a lightweight drafter $M_q$ to predict multiple candidate tokens for the next $\gamma$ steps, which are then verified in parallel by the target model $M_p$. During verification, the accepted prefix preceding the first rejected token is retained, after which $M_p$ resamples the next token from an adjusted distribution. If all candidates are accepted, $M_p$ samples a new token from its original distribution. Let $n$ denote the number of tokens produced by $M_p$ per verification step, representing the effective inference length. 

% Speculative decoding accelerates autoregressive inference by using a lightweight drafter $M_q$ to draft candidate tokens for the next $\gamma$ steps, which are then verified in parallel by the target model $M_p$. During verification, the accepted prefix before the first rejection is retained, and $M_p$ resamples the next token accordingly. If all candidates are accepted, $M_p$ samples a new token from its original distribution. The number of tokens accepted per verification step, $n$, defines the effective inference length. This process reduces sequential decoding steps, allowing multiple tokens to be generated per forward pass without sacrificing output quality.

Speculative decoding accelerates autoregressive inference by employing a lightweight drafter $M_q$ to predict candidate tokens for the next $\lambda$ steps, which are then verified in parallel by the target model $M_p$. During verification, $M_p$ accepts consecutive tokens that match its distribution and rejects the first mismatch and its follows. The accepted prefix is retained, and $M_p$ resamples the next token accordingly. If all candidates are accepted, $M_p$ samples a new token from its original distribution. The number of accepted tokens per iteration, $n$, defines the effective inference length, allowing multiple tokens to be generated per forward pass with without sacrificing output quality.

Despite employing a lightweight drafter, prior studies \cite{leviathan2023fast, chen2023accelerating} demonstrate that performance critically depends on $n$, which is influenced by the drafter's inference efficiency and its alignment with the target model.  Among existing speculative decoding frameworks for LLMs, GLIDE\cite{du2024glide} and EAGLE-2 \cite{li2024eagle2} are representative examples. GLIDE\cite{du2024glide} adopts a drafter with one self-attention and one cross-attention layer, and EAGLE-2 \cite{li2024eagle2} uses a single-layer decoder tightly aligned with the target model. Both further employ tree-based drafting to raise acceptance rates, yielding larger speedups in practice. In our experiments, we adopt the tree-based drafting structure introduced in EAGLE-2 \cite{li2024eagle2} to enable parallel candidate generation.

\begin{figure*}[!htbp]
    \centering
    \includegraphics[width=0.95\textwidth]{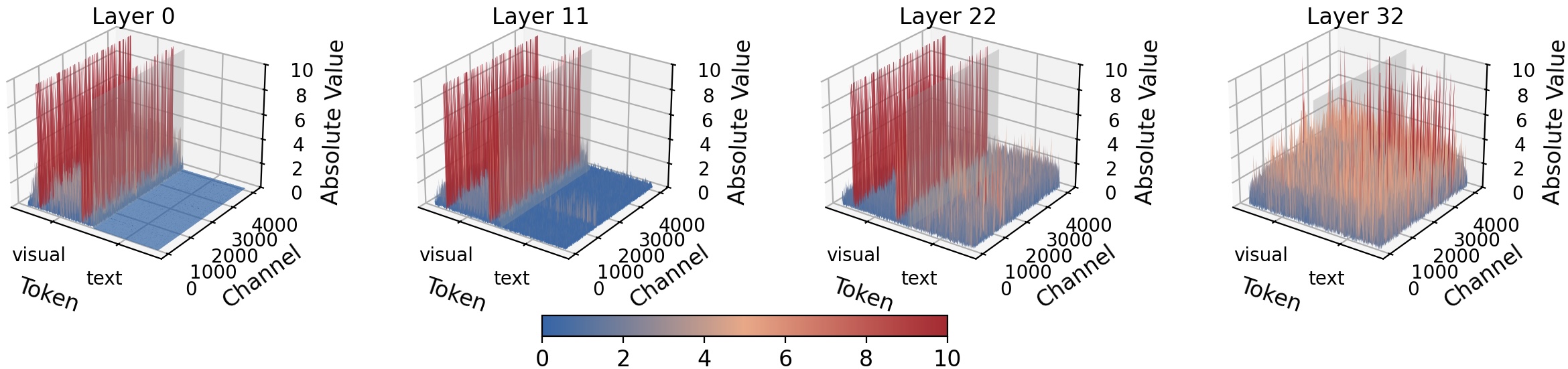}
    \caption{Visualization of hidden state magnitude distributions across layers. The gray divider marks the boundary between visual and text tokens.}
    \label{fig:selected_layers_feature}
\end{figure*}

\subsection{Large VLMs}

We focus on large VLMs \cite{wang2024emu3, liu2023llava, wang2024qwen2} that leverage pre-trained Large Language Models (LLMs) \cite{touvron2023llama, vicuna2023, team2024qwen2} as backbones, aligning visual representations with powerful language generation capabilities. Given an image-text pair, the vision encoder and multimodel projector extracts high-level visual embeddings $e_{\text{visual}} \in \mathbb{R}^{v \times d}$, which are projected into the textual embedding space, while textual inputs are embedded as $e_{\text{text}} \in \mathbb{R}^{l \times d}$. At each decoding step, the model samples the next token from $p(\cdot \mid e_{\text{visual}}, e_{\text{text}})$, enabling seamless reuse of strong LLM architectures while integrating visual semantics through lightweight projection or cross-modal alignment modules.

However, the large number of visual tokens in such models greatly inflates KV-cache size and sequence length, becoming a major bottleneck for efficient inference. Recent works alleviate this by compressing the KV-cache \cite{tu2024vl, huang2024dynamic}, compressing visual tokens \cite{zhang2025llava}, or pruning \cite{chen2024image}. To further improve inference efficiency without compromising output fidelity, we explore extending the lossless acceleration paradigm of speculative decoding to VLMs.

\section{Method}

In this section, we first delve into how visual information and visual tokens affect the drafter in speculative decoding, which has inspired us to propose the HiViS framework. Then, we will introduce the HiViS framework in details.

\subsection{Motivation}
% 分析实验：
%（1）visual features和textual features可视化；target and draft / 如何定义drafter和target每一层的相似度？使用输入输出的cosin距离，或者使用attention的KL距离？
%（2）对visual features的attention可视化，随着层数增加，对visual的attention降低
%（3）随着输出token增加，对visual的attention降低 （实验部分，可以增加实验，在不同的生成长度区间下，平均accept length的变化） / 输入的tokens逐渐减少，看看generalization能力；

% we uncover the fact that, for drafter, visual information is important, but the visual tokens is not needed.

Typically, the inputs of VLMs include visual tokens and text tokens, which are transformed into embeddings before sending to the LLM backbone as follows:
\begin{equation*}
\langle e_{\text{visual}}^1,e_{\text{visual}}^2,\dots,e_{\text{visual}}^v, e_{\text{text}}^1,e_{\text{text}}^2,\cdots,e_{\text{text}}^l \rangle,
\end{equation*}
where $e_{\text{visual}} \in \mathbb{R}^{v \times d}$ and $e_{\text{text}} \in \mathbb{R}^{l \times d}$ represent the visual and text tokens, respectively.

Current speculative decoding methods for VLMs struggle with handling visual tokens:
(1) Require the drafters to process visual tokens independently for alignment, which preserves the computational overhead of long visual sequences.
(2) The lightweight drafter, limited in capacity, often fails to capture the key visual information emphasized by the target VLM, causing semantic misalignment.
As shown in \Cref{fig:var_input}, approaches like MSD~\cite{lin2025speculative} decouple visual and textual streams but still process both, while methods like IbED~\cite{leebatch} introduce complex visual transformation before drafting, further increasing alignment difficulty. Consequently, both inefficiency and misalignment from visual tokens remain unresolved.

% On the other hand, recent studies~\cite{zhang2024cls, zhang2025vispruner} demonstrate that in VLMs \cite{liu2023llava, liu2024llavanext}, attention scores during the prefill stage guides substantial visual token pruning without notable performance degradation. This suggests that only a small subset of visual tokens carries most of the visual information, and that essential visual semantics are already embedded within the hidden representations of the VLM. Accordingly, the drafter can omit visual tokens and obtain visual semantics from the fused high-level representations produced by the target VLM.
On the other hand, recent studies~\cite{zhang2024cls, zhang2025vispruner} demonstrate that in VLMs~\cite{liu2023llava, liu2024llavanext}, prefill-stage attention guides aggressive visual token pruning with minimal performance loss. This indicates that only a small subset of visual tokens is essential and that key visual semantics are already embedded in the VLM's hidden representations. Thus, the drafter can safely omit visual tokens and rely on the fused high-level features provided by the target VLM.

These challenges raise a natural question: \emph{can the drafter avoid directly processing visual tokens while still accessing visual semantics?}
% To examine this, we conduct a variant-input experiment on EAGLE-2~\cite{li2024eagle2}, comparing full visual-text inputs with text-only inputs where all visual tokens are removed.
% As shown in Figure~\ref{fig:prove_visual}, the text-only setting outperforms the full-input setting in average acceptance length, indicating that naively reusing visual tokens for the drafter can be counterproductive.

% \begin{figure}[htbp]
%     \centering
%     \includegraphics[width=0.34\textwidth]{vision_attn_surface.png}
%     \caption*{(a) Vision Attention Surface}

%     \vspace{0.5em}

%     \includegraphics[width=0.4\textwidth]{vision_attn_mean_flat.pdf}
%     \caption*{(b) Mean Attention Curves}

%     \caption{Visualization of vision attention across layers and tokens. 
%     (a) Mean attention from post-visual tokens to visual tokens across layers. 
%     The gray plane separates the instruction region from the generated text. 
%     (b) Attention variation curves obtained by averaging over layers (left) and over tokens (right).}
%     \label{fig:vision_attn_compare}
% \end{figure}

\begin{figure}[tbp]
\centering
\begin{minipage}[b]{0.53\linewidth}
    \centering
    \subfloat[][Visual Attention Surface]{%
        \includegraphics[width=\linewidth]{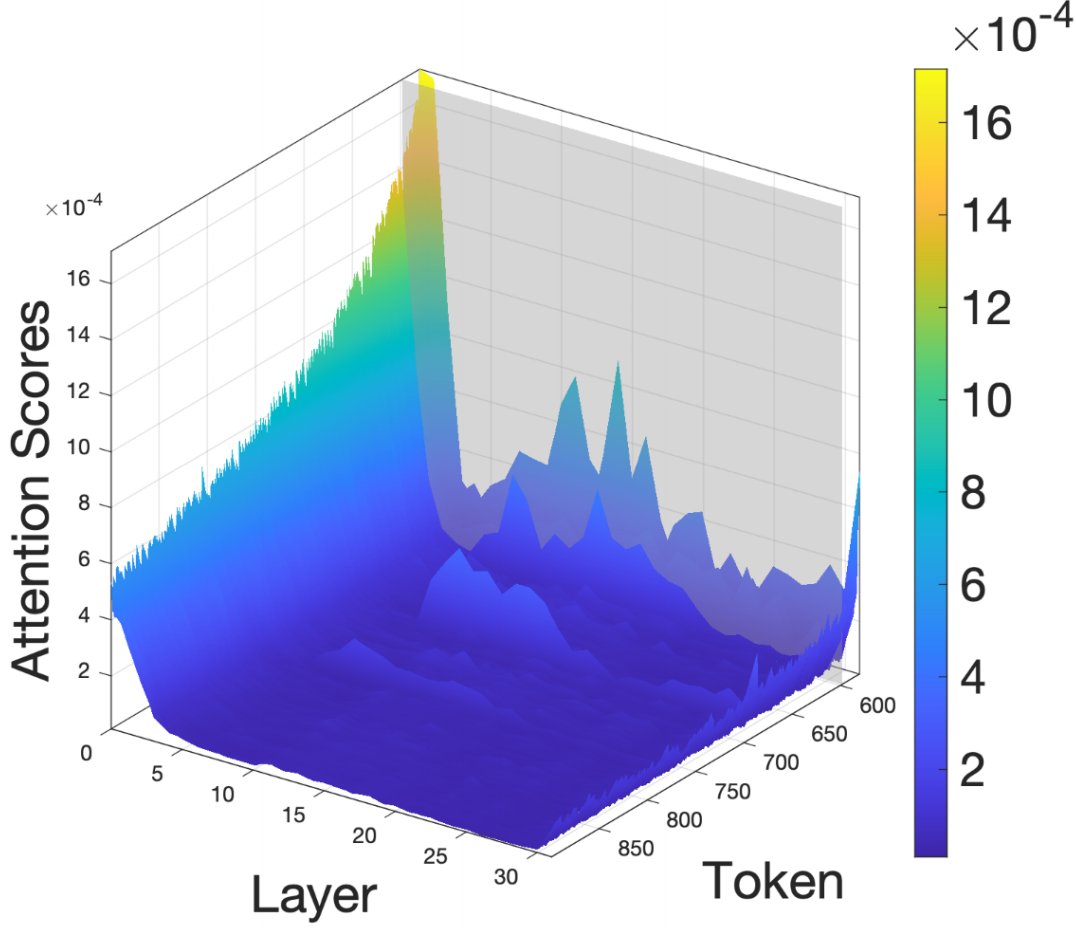}%
    }
\end{minipage}
\hfill
\begin{minipage}[b]{0.42\linewidth}
    \centering

    \subfloat[][Visual attention over layers]{%
        \includegraphics[width=\linewidth]{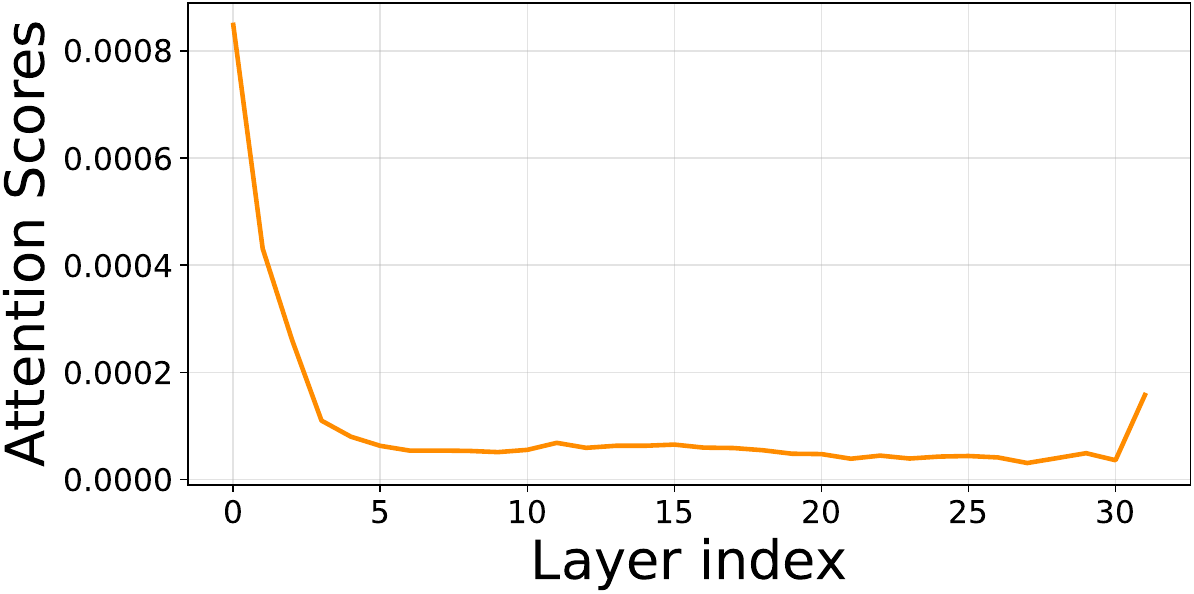}%
    }
    \vspace{0.6em}

    \subfloat[][Visual attention over tokens]{%
        \includegraphics[width=\linewidth]{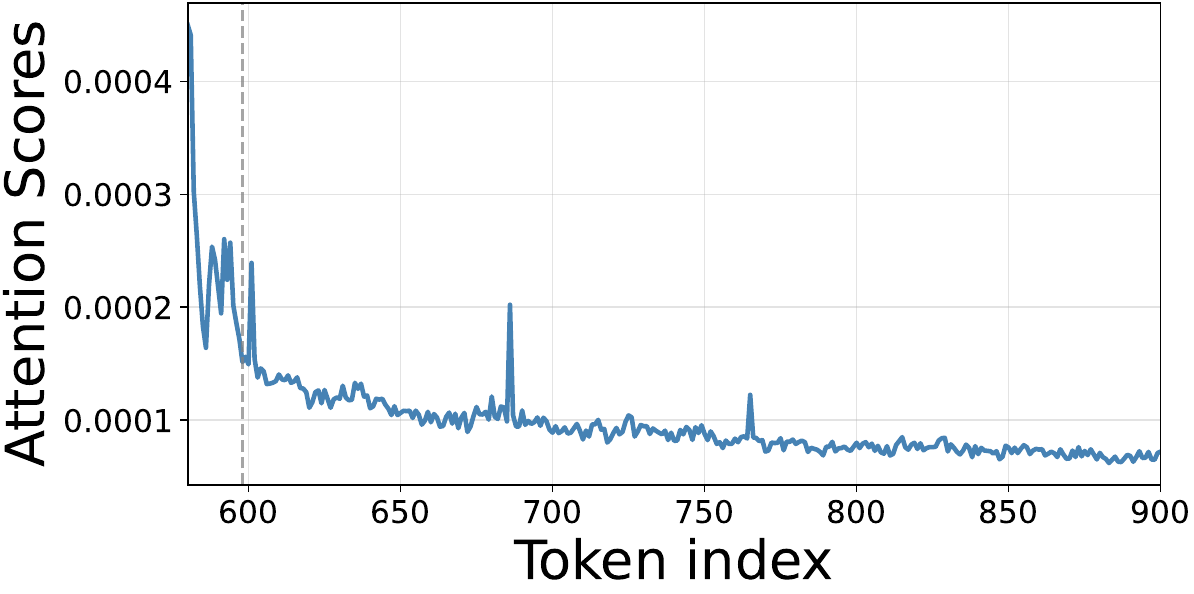}%
    }
\end{minipage}
\caption{Visualization of vision attention across layers and tokens. 
    (a) Mean attention from post-visual tokens to visual tokens across layers. 
    The gray plane separates the instruction region from the generated text. 
    (b) Attention variation curves obtained over layers. (c) Attention variation curves obtained over tokens.}
\vspace{-0.3cm}
\label{fig:vision_attn_compare}
\end{figure}

\subsection{Hiding Visual Tokens from the Drafter} \label{sec:select}

\subsubsection{Vision-to-Language Transition in Large VLMs}
To answer the above question, we conduct a layerwise and token-level analysis of hidden states and attention patterns in LLaVA-1.5-7B \cite{liu2023llava} to investigate how visual information is represented. 

% \textbf{The misalignment between visual and textual features.} 
% Modality Representation Gap
\vspace{-3mm}
\paragraph{The modality gap between visual and text tokens.} We first assess the alignment between visual and textual features in target VLMs. To this end, we visualize the feature distribution of hidden state across layers in \Cref{fig:selected_layers_feature}, with the gray plane separating visual and text tokens. From 
\Cref{fig:selected_layers_feature}, it is easy to conclude that, although large-scale modality alignment pre-training is performed, the visual and textual representation spaces are not well aligned. Specifically, the visual feature distributions exhibit much larger magnitude and variance than the textual feature distributions, especially for early layers. As a result, visual tokens can merely provide auxiliary visual cues for text tokens, rather than forming an integrated context for language modeling.

%The consequence
\vspace{-3mm}
\paragraph{The mismatch between target model's visual abstraction level and draft model's capacity.} The modality gap presents a less critical challenge in target VLM, as the target LLM relies on multiple decoder layers to progressively integrate and refine visual information within the language space. This can be confirmed from the distribution transition across layers in \Cref{fig:selected_layers_feature}. In early layers, visual tokens exhibit larger magnitudes, reflecting strong visual influence. As depth increases, visual contributions diminish while textual representations strengthen. By the final layers, the magnitudes across modalities appear approximately balanced, with a tendency toward textual dominance, reflecting a shift from vision to language space representation. The visualization of attention intensity to visual tokens across layers in \Cref{fig:vision_attn_compare} (a) and (b) also confirms that the target LLM progressively perceives visual information.

However, the draft model, which is designed with limited network capacity to ensure efficiency, lacks the necessary representational power to bridge this gap and interpret raw visual tokens directly. Feeding these raw visual tokens into the draft model would consequently impair its language modeling capacity and degrade performance. Therefore, a more effective strategy is to relieve the draft model from visual perception, while concentrating only on the language modeling ability.

%provide the draft model exclusively with textual tokens, enabling it to generate reasonable candidate sequences efficiently without being hindered by the complexities of cross-modal processing.

\vspace{-3mm}
\paragraph{The importance of text tokens gradually increases as decoding proceeds.}
% \Cref{fig:vision_attn_compare} visualizes attention from textual to visual tokens across layers and decoding steps, with the gray plane separating the instruction and generation regions. Attention to visual tokens drops sharply in early layers and steps, then stabilizes at a low level during subsequent generation. The instruction region shows higher visual attention, suggesting that visual information is primarily integrated during context establishment. These results indicate that instruction tokens encode visual information, while later decoding relies minimally on visual signals.
\Cref{fig:vision_attn_compare} visualizes textual to visual tokens attention across layers and decoding steps, with the gray plane seperating instruction and generation regions. Attention to visual tokens drops sharply in initial layers and remains low afterward. The instruction region exhibits stronger visual attention, indicating that visual information is primarily integrated during context establishment, while later decoding depends only weakly on visual tokens.

\subsubsection{Semantic Fusion and Drafter Workflow }
The analyses reveal that the target VLM's ability to perceive and understand visual tokens is enabled by its deep multi-layered architecture. However, the draft model does not have enough capacity to deal with both visual perception and contextual modeling. Thus our motivation is to remove the visual perception from the draft model.

However, we find that directly remove all visual information from the draft model is not a good choice. To further extend the observation, we conduct a variant-input experiment by varying the information provided to the drafter during training and inference. As shown in \Cref{tab:prove_visual}, 
%using visual injected text tokens, which is done by including visual information from the target VLM while removing all visual tokens (will be described in the following section), achieves the best average accept length.
including visual information from the target VLM while removing all visual tokens (here we denote this setting as visual injected text tokens, and will be described in the following section) not only eliminates redundant computation but also yields superior performance compared to using full visual-text inputs, where redundant visual information leads to suboptimal alignment. Conversely, when the drafter receives only textual inputs without any visual information, performance degrades significantly. These results demonstrate that effective visual semantics are neither captured by raw visual tokens nor recoverable from text alone, but are instead best conveyed through the fused representations provided by the target VLM. In short, for the draft model in VLM, \textbf{\emph{the visual information is indispensable, but the visual tokens are not.}} These observations support HiViS to operate without explicit visual tokens, with the help from the target VLM.

\begin{table}[t]
\centering
\small
\begin{tabular}{l c}
\toprule
\textbf{Variant input for train \& draft} & \textbf{average accept length} \\
\midrule
(a) text tokens only & 2.16 \\
(b) visual tokens \& text tokens & 3.04 \\
(c) visual injected text tokens & 3.28 \\
\bottomrule
\end{tabular}
\caption{Drafters are trained on the same multimodal dataset  and evaluated on LLaVA-Next-7B with ChartQA at temperature $0$, with three input configurations: (a) text-only input, (b) visual \& text tokens, and (c) visual injected text tokens.}
\vspace{-0.3cm}
\label{tab:prove_visual}
\end{table}

\begin{figure*}[!htbp]
    \centering
    \includegraphics[width=0.9\textwidth]{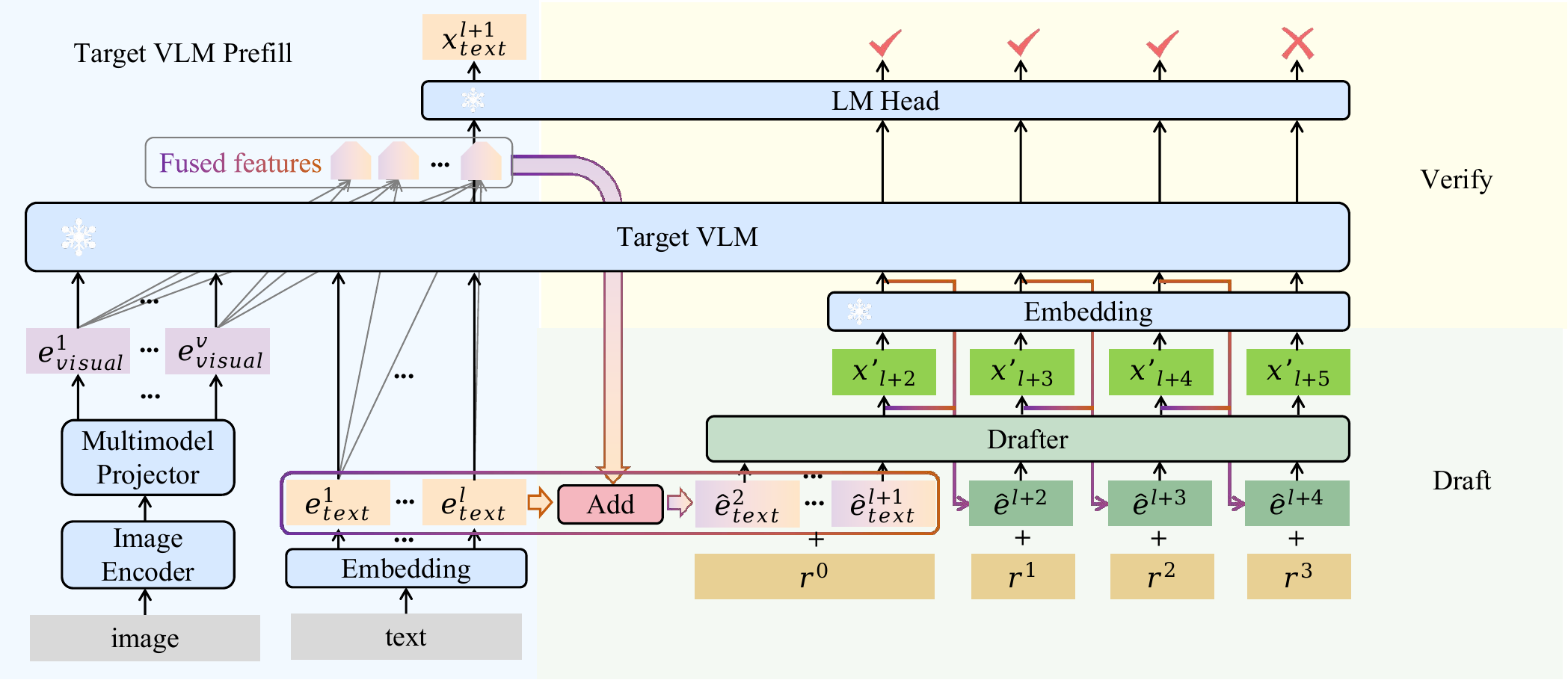}
    \caption{Overall framework of HiViS during the draft-verify process. The target VLM first performs the multimodal prefill to generate fused features $f_{\text{fused}}$ from both visual embeddings $e_{\text{visual}}$ and textual embeddings $e_{\text{text}}$ . These fused features are used to construct the visual injected text embeddings $\hat e_{\text{text}}$ that serve as inputs to the drafter. During the draft stage, the drafter generates multiple candidate, each step refined by a step-dependent bias-correction residual $r$. In the verify stage, the target VLM evaluates all candidates in parallel and accepts the consecutive correct sequence.}
\vspace{-0.3cm}
    \label{fig:inference}
\end{figure*}

% \subsection{Hiding Visual Tokens from the Drafter} \label{sec:select}

% \textcolor{blue}{Our key insight is that visual semantics can be preserved without requiring the drafter to explicitly process visual tokens. The target VLM, serving as a semantic fusion model, provides cross-modality-enriched hidden states that are linearly combined with textual embeddings to form fused inputs for the drafter, keeping the prefill sequence length identical to the textual input and thereby accelerating subsequent self-attention operations.}

% \textcolor{blue}{To enable the drafter to independently propagate and update visual-textual semantics without direct feedback from the target VLM, we introduce a dynamic multi-step training strategy that simulates the inference process. A step-dependent bias correction residual is added at each decoding step to stabilize and guide representation updates. Furthermore, a dynamic data filtering mechanism prioritizes examples within the drafter's capability while excluding persistently failed cases, improving both training efficiency and semantic consistency across extended generation.}

% The drafter consists of a single decoder layer and reuses the embedding layer and LM head of the target VLM, consistent with the design principle introduced in EAGLE \cite{li2024eagle, li2024eagle2}.
% In summary, our one-layer-decoder drafter operates on inputs with two components: (1) fused embedding with with visual guidance and textual content (2) a dynamic bias correction residual to regulate and supervise the drafting process across multiple steps.

\vspace{-3mm}
\paragraph{The overall framework.} We introduce the HiViS framework in details. Our key insight is that visual semantics can be preserved without requiring the drafter to explicitly process visual tokens. The target VLM, serving as a semantic fusion model, provides cross-modality-enriched hidden states that are linearly combined with textual embeddings to form fused inputs for the drafter, relieving the draft model from visual perception while concentrating on language modeling.

%keeping the drafter to focus on language modeling without visual perception.
%keeping the prefill sequence length identical to the textual input and thereby accelerating subsequent self-attention operations.

As illustrated in \Cref{fig:inference}, the input to the target VLM $M_p$ consists of two modalities: visual tokens and text tokens. The visual branch is obtained by passing the image through a vision encoder followed by a multimodal projector, producing a sequence of $v$ visual embeddings $e^{1}_{{visual}},\dots,e^{v}_{{visual}}$. In parallel, the text branch is formed by embedding the text tokens, yielding $e^{1}_{{text}},\dots,e^{l}_{{text}}$.

Within the target VLM $M_p$, which typically consists of multiple decoder layers (denote the number as $N$), each decoder performs self-attention over both textual and visual tokens to achieve multimodal fusion. 
At decoding step $t>v$, the input to the $n_{th}$ decoder layer is defined as $F_{n-1} = [\, f_{\text{visual}}^{1:v},\, f_{\text{text}}^{1:t-v} \,] \in \mathbb{R}^{t \times d}$, where $F_0 = [\, e_{\text{visual}}^{1:v},\, e_{\text{text}}^{1:t-v} \,]$. We employ the final-layer outputs $F_{N}^{v+1:l}$ as the fused features $f_{\text{fused}}^{1:l}$ provided to the drafter $M_q$. In addition, the drafter requires explicit textual inputs $e_{\text{text}}^{1:l} \in \mathbb{R}^{l \times d}$. We form the final visual injected text embeddings by a linear combination of the fused features and the causally shifted text embeddings $\hat e_{\text{text}}^{2:l+1} = f_{\text{fused}}^{1:l} w_1 + e_{\text{text}}^{2:l+1} w_2 \;\in\mathbb{R}^{l \times d}$, where $w_1, w_2 \in \mathbb{R}^{d \times d}$ are learnable parameters.

\begin{algorithm}[t]
\caption{HiViS: Residual-Aided Drafting and Verification Process}
\label{alg:hivis_hand}
\begin{algorithmic}[1]
\renewcommand{\algorithmicrequire}{\textbf{Inputs:}}
\Require Target VLM $M_p$, Drafter $M_q$, $e_{\text{visual}} \in \mathbb{R}^{v \times d}$, $e_{\text{text}} \in \mathbb{R}^{l \times d}$, $f_{\text{fused}}^{1:l}, e_{\text{text}}^{l+1} \leftarrow \text{Prefill}(M_p; e_{\text{visual}}^{1:v},\, e_{\text{text}}^{1:l})$
\Statex \textcolor{green!50!black}{$\triangleright$ Prefill the target VLM and obtain fused textual representations carrying visual semantics.}
% \vspace{2pt}
\State $\hat e_{\text{text}}^{2:l+1} \leftarrow f_{\text{fused}}^{1:l} w_1 + e_{\text{text}}^{2:l+1} w_2$
\Statex \textcolor{green!50!black}{$\triangleright$ Semantic Fusion: text embeddings fuse visual-aware fused feature forming visual injected text embedding.}
\State $\text{prefix} \leftarrow \hat e_{\text{text}}^{2:l} + \text{Expend}(r^0 w), \hat e^{l+1} \leftarrow \hat e_{\text{text}}^{l+1}$
\Statex \textcolor{green!50!black}{$\triangleright$ Build prefix with initial bias correction residual $r^0$.}
% \vspace{2pt}

\For{$i = 1$ \textbf{to} $\lambda$}
    \Statex \textcolor{green!50!black}{$\triangleright$ Step-wise drafting: lightweight drafter autoregressively generates fused features $f_{\text{new}}$ and tokens $e_{\text{new}}$.}
    \State $f'_{l+i} \leftarrow M_q(\text{prefix},\, \hat e^{l+1:l+i} + r^{0:i-1} w)$
    \State $x'_{l+i+1} \leftarrow \text{LM Head}(f'_{l+i})$
    \State $\hat e^{l+i+1} \leftarrow f'_{l+i} w_1 + \text{Embed}(x'_{l+i+1}) w_2$
    \Statex \textcolor{green!50!black}{$\triangleright$ Combine fused features and textual embeddings to propagate implicit visual semantics.}
\EndFor

% \vspace{2pt}
\State $x'_{l+2:l+1+n} \leftarrow \text{Parallel Verify}(M_p,\, x'_{l+2:l+1+\lambda})$
\Statex \textcolor{green!50!black}{$\triangleright$ Parallel verification by target VLM.}
% \vspace{2pt}
\State \Return $e_{\text{text}} \leftarrow e_{\text{text}}^{1:l} \oplus \text{Embed}(x'_{l+1:l+1+n})$
% \Statex \textcolor{green!50!black}{$\triangleright$ Append accepted embeddings with residual for the next decoding iteration.}
\end{algorithmic}
\end{algorithm}

\Cref{alg:hivis_hand} shows a detailed utilization of the visual injected text embedding within the draft-verify cycle. During the draft stage, since the drafter cannot access the target VLM's fused features, at draft step $i$ it substitutes its own hidden state from the previous decoder layer, $f'_{l+i}$, to construct the current representation $\hat e^{l+i+1}$. The input is then formed as $\hat e^{l+i+1} + r^{i} w$, where $r$ is the learnable bias correction residual specifically tailored for the draft phase to alleviate the accumulation of error arising from the visual injected text embedding over time, with $w\!\in\!\mathbb{R}^{d\times d}$. The detailed design and training of this residual mechanism are described in the following section.

\subsection{Time-step-aware Residual Aligned Training}

The HiViS framework can overcome the visual token hurdle in the draft model, however, mismatch still exists between the drafter and the target model. The main reason for this mismatch is that during drafting stage, the drafter cannot obtain the target VLM's latest visual-textual semantic updates. We will provide a detailed elaboration of the problem and the corresponding solution.

% 获得不了真实的target feature，所以引入residual
% 由于draft采用拟合的target feature，随着序列长度增加会跑偏，所以早停

% We have three training objectives: (1) to align the drafter's token predictions with those of the target VLM, ensuring alignment in output distribution. (2) To align the drafter's hidden states with the corresponding fused feature of the target VLM. (3) To learn step-dependent bias-correction residuals at each simulated forward step. Thereby enabling the drafter to learn the progressive update of visual semantics that occurs during inference.
\vspace{-3mm}
\paragraph{Step-dependent bias-correction residuals.}
%HiViS conditions the drafter on visual injected text embeddings whose fused-feature component is supplied by the target VLM. This setup brings two considerations: (1) the visual injected text embeddings must be adapted to the drafter's shallow architecture, and (2) during independent inference the drafter cannot obtain the target VLM's latest visual-textual semantic updates. To address both issues, we design a time-step aligned training strategy with step-dependent bias-correction residuals, which are linearly added to the visual injected text embeddings at each step. These residuals progressively adjust the visual injected text embeddings to match the drafter and support stable semantic propagation over independent drafting steps. Furthermore, a dynamic top-k filtering strategy keeps within-capacity examples while removing consistently failed ones, enhancing training efficiency and semantic consistency across extended generation.

\begin{figure}[!t]
    \centering
    \includegraphics[width=0.45\textwidth]{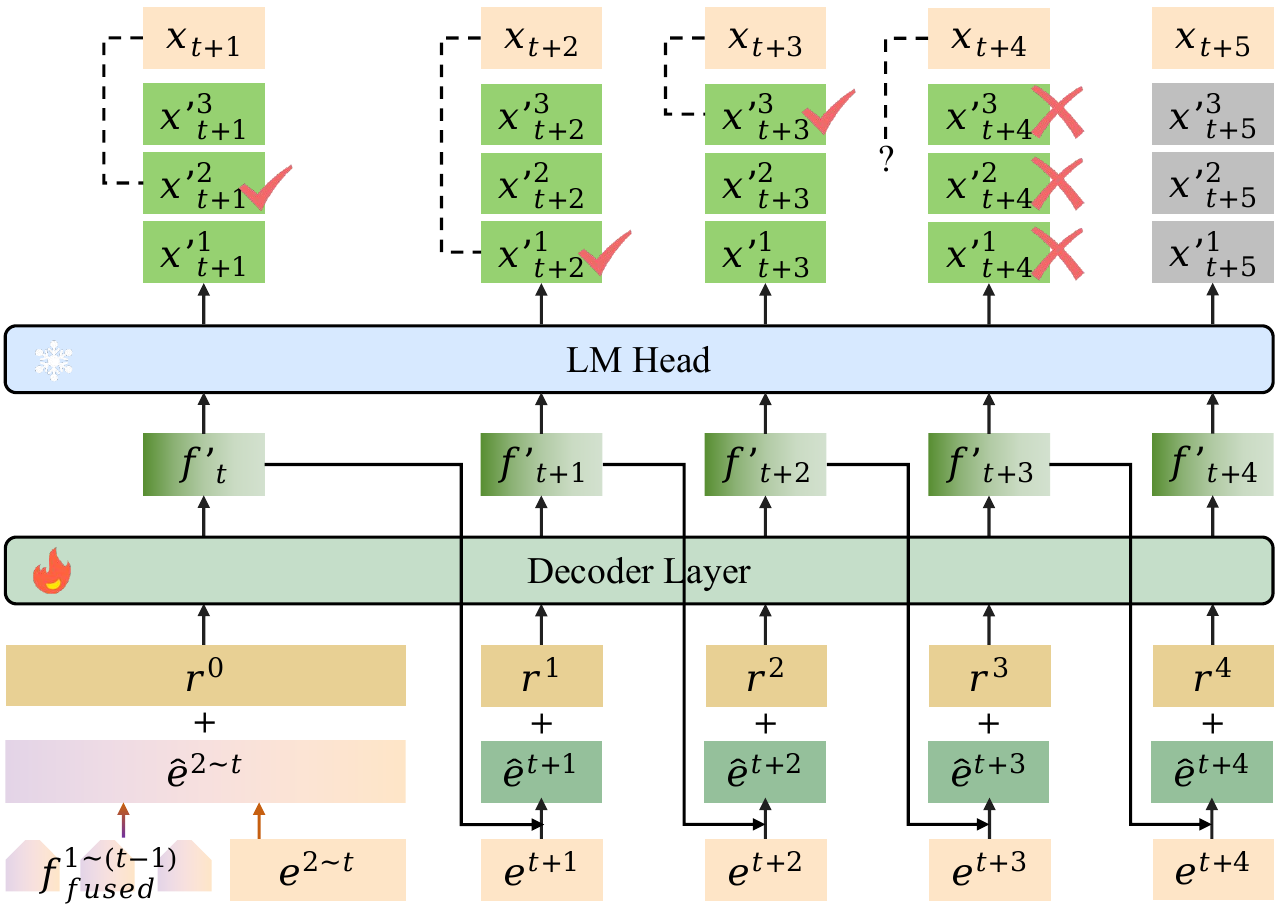}
    \caption{Training architecture of HiViS. $e$ are input embeddings, $f_{\text{fused}}$ and $x$ are fused feature and tokens from the target VLM, while $f'$ denotes the drafter-predicted hidden states and $x'^k$ the corresponding top-k token candidates drawn from its output distribution. Correct predictions are marked with $\checkmark$, errors with $\times$, and gray tokens denote discarded candidates.}
    \vspace{-0.3cm}
    \label{fig:training}
\end{figure}

HiViS conditions the drafter on visual injected text embeddings whose fused-feature component is supplied by the target VLM. However, this information is not available during drafting. Instead, the hidden state from the previous step of the draft model is used, i.e., the $f'_{t}$ in \Cref{fig:training}, which causes the mismatch between the drafter and the target model. To fill this gap, we design a step-dependent bias-correction residuals (the $r^i$ in \Cref{fig:training}), which are linearly added to the visual injected text embeddings at each draft decoding step. These residuals progressively adjust the visual injected text embeddings to support stable semantic propagation over independent drafting steps. 

To train these step-dependent bias-correction residual terms, HiViS performs $N$ step inference simulation to mimic the drafter's independent decoding process, as illustrated in in \Cref{fig:training} (with $N=5$) \cite{zhang2024learning}. At the starting timestep $t$ for each drafting round, the drafter receives the embeddings up to $t$, denoted as $[e^1, e^2, \ldots, e^t]$. The initial visual injected text embedding is constructed as $\hat e^{2:t} = f_{\text{fused}}^{1:(t-1)} w_1 + e^{2:t} w_2$, where $f_{\text{fused}}^{1:(t-1)}$ represents the fused hidden states provided by the target VLM. At the first simulation step, the initial residual $r^{0}$ is applied directly to all visual injected text embeddings to better align with the drafter's processing. Each subsequent simulation step $i$ has its own residual $r^{i-1}$, which is linearly combined with the input embedding before being fed into the drafter. These residuals progressively refine the representations and compensate for deviations that accumulate across multiple simulation steps. As shown in \Cref{fig:training}, in the first forward pass ($i=1$), the drafter predicts its hidden state as:
\begin{equation}
%f'_{t} = M_q\!\left( \hat e^2 + r^0 w, \cdots, \hat e^t + r^0 w \right).
f'_{t} = M_q\!\left( \hat e^{2:t} + r^0 w \right).
\end{equation}
Subsequently, the $i_{th}$ forward pass ($1 < i \leq N$) can be generalized as:
% \begin{align}
%     \centering
%     \small
%     {f'}_{t+i} = &M_q(\hat e^{2:t} + r^0 w, \hat e^{t+1} + r^1 w,\cdots, \hat e^{t+i-1} + r^{i-1} w).
% \end{align}
\begin{equation}
    \centering
\resizebox{0.9\hsize}{!}{$
    {f'}_{t+i} = M_q(\hat e^{2:t} + r^0 w, \hat e^{t+1} + r^1 w,\cdots, \hat e^{t+i-1} + r^{i-1} w).
$}
\end{equation}

%To address both issues, we design a time-step aligned training strategy with step-dependent bias-correction residuals, which are linearly added to the visual injected text embeddings at each step. These residuals progressively adjust the visual injected text embeddings to match the drafter and support stable semantic propagation over independent drafting steps. 

%We have two training objectives: (1) To align the drafter's output distribution with that of the target VLM, ensuring consistency in token prediction. (2) To learn step-dependent bias-correction residuals at each simulated forward step. Thereby enabling the drafter to capture the progressive update of visual-textual semantics during inference.

\vspace{-3mm}
\paragraph{Two-stage training with early stopping} Furthermore, we propose a dynamic top-K filtering strategy with early stopping to keep within-capacity examples while removing consistently failed hard token samples, enhancing training efficiency and semantic consistency across extended generation. Specifically, our training contains two stages, at the stage-1, we only train the drafter with one step, and at the stage-2, we train it for $N>1$ steps. 
In this way, the stage-1 training establishes a solid foundation for the drafter's predictive capability, while the stage-2 both simulates the draft token tree construction process of actual drafting and performs dynamic data filtering.

For stage-2, inspired by knowledge distillation \cite{xu2024speculative}, the drafter outputs a probability distribution at each step, from which the top-$K$ tokens are selected as candidates. We then apply a dynamic filtering criterion:
\begin{equation}
    \centering
    x_{t+i} \stackrel{?}{\in} \{ {x'}_{t+i}^1, \cdots, {x'}_{t+i}^k, \cdots, {x'}_{t+i}^K \}
\end{equation}
% where ${x'}_{t+i}^k$ denotes the token with the $k_{th}$ highest probability predicted by the drafter.
% If the target VLM generated ground-truth token $x_{t+i}$ falls within the drafter's top-$K$ predictions, the example at $i_{th}$ forward pass (also denoted as timestep $t+i-1$) is retained for optimization. Otherwise, the simulation for this instance is terminated early and discarded, preventing the model from wasting capacity on consistently mispredicted sequences. For instance, as shown in \Cref{fig:training} with $N=5$, the process stops at the $4_{th}$ forward pass because the candidate tokens fail to cover the target VLM's prediction, and thus the $5_{th}$ forward pass is not included in training.
where ${x'}_{t+i}^k$ denotes the token with the $k_{th}$ highest probability predicted by the drafter. If the target VLM generated ground-truth token $x_{t+i}$ appears in the drafter's top-$K$ predictions, the example at $i_{th}$ forward pass (timestep $t+i-1$) is kept for optimization. Otherwise, the simulation is terminated early and discarded to avoid wasting effort on consistently mispredicted sequences. As shown in \Cref{fig:training} with $N=5$, the process stops at the $4_{th}$ forward pass for the candidate tokens no longer cover the target VLM's prediction, and thus the $5_{th}$ forward pass is excluded from training.

\vspace{-3mm}
\paragraph{Training loss.}
    The overall loss consists of three components, weighted as:
    \begin{equation}
    L \;=\; L_{\text{fus}} \;+\;\beta \, L_{\text{cls}} \;+\; \gamma\, L_{\text{topK}},
    \end{equation}
    where 
    \[
    \begin{aligned}
    L_{\text{fus}} &= \mathrm{Smooth}\text{-}\mathrm{L1}\!\left(f_{\text{fused}}^{t},\, f_{t}'\right),\\[3pt]
    L_{\text{cls}} &= \mathrm{CrossEntropy}\!\left(p_t,\, q_t\right),\\[3pt]
    L_{\text{topK}} &= - \!\!\sum_{x \in \Omega_t} q_t(x)\,\log p_t(x).
    \end{aligned}
    \]
    Here, \(f_{\text{fused}}^{t}\) denotes the target VLM's fused feature at step \(t\) and \(f_t'\) the drafter's hidden state, \(p_t\) and \(q_t\) are the predicted probability distributions of target VLM and drafter, \(\Omega_t\) is the set of top-\(K\) tokens under the target at step \(t\). 
    Accordingly, \(L_{\text{fus}}\) supervises representation alignment supporting propagation of visual-textual semantics, 
    \(L_{\text{cls}}\) enforces distributional alignment by matching next-token logits, and \(L_{\text{topK}}\) emphasizes the ranking loss \cite{zhang2024learning} for augmenting the alignment between the top-ranked tokens.

% \subsection{Mixed Modality Dataset}

%     \textcolor{blue}{To adapt the lightweight drafter to text-only inputs, we adopt consistent preprocessing during multimodal training. Instead of storing full visual token sequences, only textual embeddings and the corresponding last-layer hidden states from the target VLM are retained, alleviating training overhead while preserving multimodal alignment, for example reducing processed data storage by up to $90\%$ for LLaVA-Next. Since many multimodal samples contain single-token answers unsuitable for multi-step training, we filter out entries with answers shorter than five words. }

%     \textcolor{blue}{Since the attention to visual tokens in VLMs gradually diminishes during text generation and most multimodal datasets contain relatively short textual segments, we further augment the training corpus with additional text-only dataset to enhance the drafter's linguistic fluency.}

\begin{table*}[t]
\centering
\scriptsize
% \small
\caption{Speedup ratio ($SR$) and average acceptance length ($\tau$) across benchmarks.}
\setlength{\tabcolsep}{2pt}
\begin{tabular}{@{}l l cc cc cc cc cc cc cc cc cc@{}}
\toprule
\multirow{2}{*}{Model} & \multirow{2}{*}{Methods} & \multicolumn{2}{c}{ChartQA}  & \multicolumn{2}{c}{VQAv2}  & \multicolumn{2}{c}{ScienceQA}  & \multicolumn{2}{c}{TextVQA}  & \multicolumn{2}{c}{MME}  & \multicolumn{2}{c}{MMVet} & \multicolumn{2}{c}{SEED-Bench} & \multicolumn{2}{c}{GQA} & \multicolumn{2}{c}{Avg} \\
&  & $SR$ & $\tau$  & $SR$ & $\tau$  & $SR$ & $\tau$  & $SR$ & $\tau$  & $SR$ & $\tau$  & $SR$ & $\tau$  & $SR$ & $\tau$ & $SR$ & $\tau$ & $SR$ & $\tau$ \\
\midrule
\multicolumn{20}{c}{T = 0} \\
\midrule
\multirow{3}{*}{LLaVA-1.5-7B} & EAGLE-2 & 1.56$\times$ & 3.06 & 1.95$\times$ & 4.62 & 1.44$\times$ & 2.97 & 1.88$\times$ & 3.38 & 1.77$\times$ & 3.72 & 1.89$\times$ & 3.71 & 2.00$\times$ & 4.64 & 1.83$\times$ & 4.39 & 1.79$\times$ & 3.81 \\
 & MSD & 2.06$\times$ & 4.53 & 2.01$\times$ & 5.06 & 2.03$\times$ & 4.27 & 2.09$\times$ & 4.22 & 2.02$\times$ & 4.74 & 2.06$\times$ & 4.38 & 2.05$\times$ & 4.97 & 1.99$\times$ & 5.16 & 2.04$\times$ & 4.67 \\
  & HiViS & \textbf{2.37$\times$} & \textbf{5.13} & \textbf{2.28$\times$} & \textbf{5.78} & \textbf{2.26$\times$} & \textbf{5.12} & \textbf{2.55$\times$} & \textbf{4.95} & \textbf{2.44$\times$} & \textbf{5.44} & \textbf{2.42$\times$} & \textbf{5.03} & \textbf{2.29$\times$} & \textbf{5.56} & \textbf{2.25$\times$} & \textbf{5.95} & \textbf{2.36$\times$} & \textbf{5.37} \\
\midrule
\multirow{3}{*}{LLaVA-1.5-13B} & EAGLE-2 & 1.73$\times$ & 3.18 & 2.05$\times$ & 4.78 & 1.57$\times$ & 2.10 & 1.95$\times$ & 3.31 & 1.84$\times$ & 3.54 & 2.04$\times$ & 3.75 & 2.07$\times$ & 4.62 & 1.92$\times$ & 4.50 & 1.90$\times$ & 3.72 \\
 & MSD & 2.15$\times$ & 4.28 & 2.07$\times$ & 4.59 & 2.02$\times$ & 3.87 & 2.12$\times$ & 3.92 & 2.07$\times$ & 3.82 & 2.14$\times$ & 4.08 & 2.11$\times$ & 4.70 & 2.00$\times$ & 4.58 & 2.09$\times$ & 4.23 \\
 & HiViS & \textbf{2.43$\times$} & \textbf{5.24} & \textbf{2.20$\times$} & \textbf{5.44} & \textbf{2.17$\times$} & \textbf{4.76} & \textbf{2.50$\times$} & \textbf{4.69} & \textbf{2.35$\times$} & \textbf{5.04} & \textbf{2.52$\times$} & \textbf{5.05} & \textbf{2.28$\times$} & \textbf{5.46} & \textbf{2.15$\times$} & \textbf{5.51} & \textbf{2.33$\times$} & \textbf{5.15} \\
\midrule
\multirow{4}{*}{LLaVA-Next-7B} & EAGLE-2 & 1.33$\times$ & 3.04 & 1.59$\times$ & 4.52 & 1.47$\times$ & 3.00 & 1.43$\times$ & 3.04 & 1.43$\times$ & 3.35 & 1.42$\times$ & 2.81 & 1.36$\times$  & 4.33  & 1.39$\times$ & 4.36  & 1.43$\times$  & 3.56 \\
 & Dream & 1.28$\times$  & 3.62  & 1.34$\times$  & 3.21  & 1.39$\times$  & 3.65  & 1.06$\times$  & 2.72  & 1.20$\times$  & 3.30  & 1.16$\times$  & 2.82  & 1.36$\times$  & 4.35  & 1.32$\times$  & 4.28  & 1.26$\times$  & 3.49 \\
 & ViSpec & 1.66$\times$  & 4.36  & 1.53$\times$  & 4.17  & 1.78$\times$  & 4.19  & 1.99$\times$  & 4.46  & 1.59$\times$  & 4.03  & 1.70$\times$  & 4.17  & 1.58$\times$  & 4.33  & 1.38$\times$  & 4.18  & 1.65$\times$  & 4.24 \\
 & HiViS & \textbf{1.87$\times$} & \textbf{4.96} & \textbf{1.79$\times$} & \textbf{5.47} & \textbf{2.11$\times$} & \textbf{4.98} & \textbf{2.30$\times$} & \textbf{5.06} & \textbf{1.81$\times$} & \textbf{4.68} & \textbf{1.97$\times$} & \textbf{4.85} & \textbf{1.79$\times$} & \textbf{5.16} & \textbf{1.59$\times$} & \textbf{5.35} & \textbf{1.90$\times$} & \textbf{5.06} \\
\midrule
\multirow{3}{*}{LLaVA-Next-13B} & EAGLE-2 & 1.43$\times$ & 3.19 & 1.63$\times$ & 4.55 & 1.56$\times$ & 3.06 & 1.52$\times$ & 2.98 & 1.53$\times$ & 3.39 & 1.54$\times$ & 3.23 &1.28$\times$  & 4.33  & 1.31$\times$ & 4.38  & 1.48$\times$  & 3.64 \\
 & ViSpec & 1.76$\times$  & 4.47  & 1.61$\times$  & 4.25  & 1.87$\times$  & 4.03  & 2.03$\times$  & 4.25  & 1.72$\times$  & 4.10  & 1.76$\times$  & 3.96  & 1.74$\times$  & 4.62  & 1.34$\times$  & 4.26  & 1.73$\times$  & 4.24 \\
 & HiViS & \textbf{1.88$\times$} & \textbf{5.05} & \textbf{1.79$\times$} & \textbf{5.48} & \textbf{2.14$\times$} & \textbf{4.80} & \textbf{2.33$\times$} & \textbf{4.90} & \textbf{1.89$\times$} & \textbf{4.70} & \textbf{1.97$\times$} & \textbf{4.67} & \textbf{1.85$\times$} & \textbf{5.25} & \textbf{1.43$\times$} & \textbf{5.26} & \textbf{1.91$\times$} & \textbf{5.01} \\
\midrule
\multirow{3}{*}{Qwen2.5-VL-7B} & EAGLE-2 & 1.33$\times$  & 2.49  & 2.09$\times$  & 4.40  & 1.41$\times$  & 2.62  & 2.31$\times$  & 2.78  & 1.60$\times$  & 2.96  & 1.52$\times$  & 2.57 & 1.80$\times$  & 3.79  & 1.96$\times$  & 4.14  & 1.75$\times$  & 3.22 \\
& ViSpec & 1.81$\times$  & 3.90  & 1.90$\times$  & 3.73  & 1.86$\times$  & 3.71  & 2.87$\times$  & 3.60  & 1.75$\times$  & 3.46  & 1.75$\times$  & 3.44  & 1.69$\times$  & 3.68  & 1.92$\times$  & 3.87  & 1.94$\times$  & 3.67 \\
 % & HiViS (qwen) & 2.01$\times$ & 4.41 & 2.28$\times$ & 4.74 & 2.08$\times$ & 4.11 & 3.12$\times$ & 3.86 & 1.97$\times$ & 3.85 & 1.98$\times$ & 3.85 & 2.24$\times$ & 4.14 \\
 & HiViS & \textbf{2.04$\times$} & \textbf{4.56} & \textbf{2.28$\times$} & \textbf{4.92} & \textbf{2.13$\times$} & \textbf{4.34} & \textbf{3.15$\times$} & \textbf{3.95} & \textbf{1.98$\times$} & \textbf{3.98} & \textbf{2.00$\times$} & \textbf{4.01} & \textbf{2.00$\times$} & \textbf{4.54} & \textbf{2.21$\times$} & \textbf{4.84} & \textbf{2.22$\times$} & \textbf{4.39} \\
\midrule
\multicolumn{20}{c}{T = 1} \\
\midrule
\multirow{3}{*}{LLaVA-1.5-7B} & EAGLE-2 & 1.42$\times$ & 2.66 & 1.77$\times$ & 3.68 & 1.29$\times$ & 2.58 & 1.38$\times$ & 2.49 & 1.49$\times$ & 2.85 & 1.53$\times$ & 2.82 & 1.77$\times$  & 3.51  & 1.71$\times$  & 3.64  & 1.55$\times$  & 3.03 \\
 & MSD & 1.70$\times$ & 3.29 & 1.79$\times$ & 3.86 & 1.59$\times$ & 3.27 & 1.51$\times$ & 2.96 & 1.65$\times$ & 3.36 & 1.66$\times$ & 3.23 & 1.76$\times$  & 3.84  & 1.82$\times$  & 4.00  & 1.69$\times$  & 3.48 \\
 & HiViS & \textbf{1.95$\times$} & \textbf{3.60} & \textbf{1.97$\times$} & \textbf{3.99} & \textbf{1.84$\times$} & \textbf{3.63} & \textbf{1.71$\times$} & \textbf{3.32} & \textbf{1.89$\times$} & \textbf{3.61} & \textbf{1.90$\times$} & \textbf{3.48} & \textbf{1.95$\times$} & \textbf{3.92} & \textbf{2.00$\times$} & \textbf{4.34} & \textbf{1.90$\times$} & \textbf{3.74} \\
\midrule
\multirow{3}{*}{LLaVA-1.5-13B} & EAGLE-2 & 1.57$\times$ & 2.80 & 1.83$\times$ & 3.74 & 1.52$\times$ & 2.65 & 1.58$\times$ & 2.65 & 1.66$\times$ & 2.91 & 1.69$\times$ & 2.91 & 1.85$\times$  & 3.71  & 1.86$\times$  & 3.74  & 1.70$\times$  & 3.14 \\
 & MSD & 1.86$\times$ & 3.40 & 1.92$\times$ & 3.72 & 1.71$\times$ & 3.18 & 1.66$\times$ & 3.00 & 1.76$\times$ & 3.24 & 1.80$\times$ & 3.15 & 1.90$\times$  & 3.71  & 1.85$\times$  & 3.86  & 1.81$\times$  & 3.41 \\
 & HiViS & \textbf{2.04$\times$} & \textbf{3.70} & \textbf{1.96$\times$} & \textbf{4.18} & \textbf{1.92$\times$} & \textbf{3.49} & \textbf{1.91$\times$} & \textbf{3.12} & \textbf{2.02$\times$} & \textbf{3.78} & \textbf{2.00$\times$} & \textbf{3.58} & \textbf{2.02$\times$} & \textbf{4.07} & \textbf{1.98$\times$} & \textbf{4.11} & \textbf{1.98$\times$} & \textbf{3.75} \\
\midrule
\multirow{4}{*}{LLaVA-Next-7B} & EAGLE-2 & 1.18$\times$ & 2.63 & 1.43$\times$ & 3.46 & 1.20$\times$ & 2.49 & 1.07$\times$ & 2.30 & 1.26$\times$ & 2.63 & 1.23$\times$ & 2.65 & 1.21$\times$  & 3.23  & 1.24$\times$  & 3.42  & 1.23$\times$  & 2.85 \\
 & Dream & 1.18$\times$ & 3.31 & 1.25$\times$ & 3.84 & 1.33$\times$ & 3.34 & 0.96$\times$ & 2.51 & 1.14$\times$ & 3.1 & 1.11$\times$  & 2.87  & 1.26$\times$  & 3.01  & 1.22$\times$  & \textbf{4.03} & 1.18$\times$  & 3.25 \\
 & ViSpec & 1.46$\times$ & 3.49 & 1.42$\times$ & 3.37 & 1.40$\times$ & 3.18 & 1.35$\times$ & 3.04 & 1.36$\times$ & 3.15 & 1.40$\times$  & 3.11  & 1.31$\times$  & 3.40  & 1.25$\times$  & 3.38  & 1.37$\times$  & 3.27 \\
 & HiViS & \textbf{1.60$\times$} & \textbf{3.86} & \textbf{1.56$\times$} & \textbf{4.07} & \textbf{1.61$\times$} & \textbf{3.54} & \textbf{1.48$\times$} & \textbf{3.21} & \textbf{1.53$\times$} & \textbf{3.46} & \textbf{1.54$\times$} & \textbf{3.47} & \textbf{1.47$\times$} & \textbf{3.78} & \textbf{1.37$\times$} & 3.81 & \textbf{1.52$\times$} & \textbf{3.65} \\
\midrule
\multirow{2}{*}{LLaVA-Next-13B} & EAGLE-2 & 1.27$\times$ & 2.71 & 1.51$\times$ & 3.47 & 1.33$\times$ & 2.63 & 1.26$\times$ & 2.46 & 1.41$\times$ & 2.87 & 1.35$\times$ & 2.72 & 1.21$\times$  & 3.29  & 1.25$\times$  & 3.52  & 1.32$\times$  & 2.96 \\
 & ViSpec & 1.57$\times$  & 3.58  & 1.53$\times$  & 3.44  & 1.64$\times$  & 3.21  & 1.52$\times$  & 3.21  & 1.59$\times$  & 3.30  & 1.53$\times$  & 3.24  & 1.49$\times$  & 3.50  & 1.24$\times$  & 3.49  & 1.51$\times$  & 3.37 \\
 & HiViS & \textbf{1.76$\times$} & \textbf{3.82} & \textbf{1.74$\times$} & \textbf{3.99} & \textbf{1.75$\times$} & \textbf{3.48} & \textbf{1.68$\times$} & \textbf{3.37} & \textbf{1.68$\times$} & \textbf{3.60} & \textbf{1.66$\times$} & \textbf{3.54} & \textbf{1.61$\times$} & \textbf{3.85} & \textbf{1.25$\times$} & \textbf{3.89} & \textbf{1.64$\times$} & \textbf{3.69} \\
\midrule
\multirow{3}{*}{Qwen2.5-VL-7B} & EAGLE-2 & 1.28$\times$  & 2.53  & 1.72$\times$  & 3.34  & 1.33$\times$  & 2.50  & 1.87$\times$  & 2.37  & 1.42$\times$  & 2.63  & 1.31$\times$  & 2.35  & 1.64$\times$  & 3.24  & 1.71$\times$  & 3.33  & 1.54$\times$  & 2.79 \\
 & ViSpec & 1.72$\times$  & 3.88  & 1.63$\times$  & 3.10  & 1.72$\times$  & 3.45  & 2.27$\times$  & 2.96  & 1.51$\times$  & 2.97  & 1.54$\times$  & 3.06  & 1.55$\times$  & 3.41  & 1.62$\times$  & 3.20  & 1.70$\times$  & 3.25 \\
 % & HiViS (qwen) & 1.89$\times$ & 4.27 & 1.84$\times$ & 3.69 & 1.88$\times$ & 3.78 & 2.42$\times$ & 3.06 & 1.70$\times$ & 3.25 & 1.67$\times$ & 3.23 & 1.90$\times$ & 3.55 \\
 & HiViS & \textbf{1.89$\times$} & \textbf{4.38} & \textbf{1.86$\times$} & \textbf{3.63} & \textbf{1.97$\times$} & \textbf{3.97} & \textbf{2.39$\times$} & \textbf{3.08} & \textbf{1.70$\times$} & \textbf{3.27} & \textbf{1.67$\times$} & \textbf{3.40} & \textbf{1.79$\times$} & \textbf{3.72} & \textbf{1.86$\times$} & \textbf{3.84} & \textbf{1.89$\times$} & \textbf{3.66} \\
\midrule
\bottomrule
\end{tabular}
\label{tab:depth5_result}
\end{table*}

\section{Experiment}

\textbf{Models and Tasks: }
    % To evaluate the generality and effectiveness of the proposed approach across different task types, we conduct experiments on three representative open-source VLMs: LLaVA-1.5 (7B and 13B) \cite{liu2023llava}, LLaVA-Next (7B and 13B) \cite{liu2024llavanext} and Qwen2.5VL-7B \cite{bai2025qwen2}. We conduct experiments on common visual question answering benchmarks, including ChartQA \cite{masry2022chartqa}, VQAv2 \cite{goyal2017making}, ScienceQA \cite{lu2022learn}, TextVQA \cite{singh2019towards}, MME \cite{fu2025mme}, MM-Vet \cite{yu2023mm}, SEED-Bench\cite{li2023seed} and GQA\cite{hudson2019gqa}, all of which assess image understanding and reasoning capabilities. We evaluate each task on a subset of 80 samples.
    To assess the effectiveness and generality of HiViS, we evaluate three representative open-source VLMs: LLaVA-1.5 (7B, 13B) \cite{liu2023llava}, LLaVA-Next (7B, 13B) \cite{liu2024llavanext}, and Qwen2.5-VL-7B \cite{bai2025qwen2}. Experiments are conducted on standard VQA benchmarks, including ChartQA \cite{masry2022chartqa}, VQAv2 \cite{goyal2017making}, ScienceQA \cite{lu2022learn}, TextVQA \cite{singh2019towards}, MME \cite{fu2025mme}, MM-Vet \cite{yu2023mm}, SEED-Bench \cite{li2023seed}, and GQA \cite{hudson2019gqa}, covering diverse image understanding and reasoning tasks. Each benchmark is evaluated on an 80-sample subset.

\textbf{Baselines: }
    % We compare HiViS with four baselines, all evaluated using their publicly released model weights.
    % (1) EAGLE-2 \cite{li2024eagle2} serves as our foundation, adapted to the VLM setting by incorporating both visual and textual tokens and retraining on the same multimodal dataset.
    % (2) MSD \cite{lin2025speculative} extends speculative decoding to VLMs with a single-layer decoder drafter that decouples visual and textual processing but still relies on full multimodal inputs.
    % (3) ViSpec \cite{kang2025vispec} represents the latest speculative decoding framework for VLMs, where a single-layer drafter explicitly processes and compresses visual tokens during inference.
    % (4) Dream \cite{hu2025dream} adopts a two-layer decoder with one cross-attention module as the drafter and applies visual token pruning to accelerate inference.
    We compare HiViS with four baselines.
    EAGLE-2 \cite{li2024eagle2}, a one-decoder-layer speculative decoding framework for LLMs, is reimplemented and retrained on the multimodal dataset.
    For VLM-specific baselines, we evaluate MSD \cite{lin2025speculative}, ViSpec \cite{kang2025vispec}, and Dream \cite{hu2025dream} using their publicly released model weights.
    MSD employs a single-layer decoder drafter that decouples visual and textual processing but still relies on full multimodal inputs.
    ViSpec uses a single-layer drafter that explicitly compresses visual tokens during inference, while Dream adopts a two-layer decoder with one cross-attention module and applies visual token pruning to accelerate generation.

\textbf{Metrics: }
    % As speculative decoding serves as the core mechanism in this work, the evaluation follows standard protocols and focuses on inference efficiency. Two primary metrics are employed:
    As speculative decoding is central to this work, evaluation follows standard protocols and focuses on inference efficiency, using two primary metrics:
    \begin{itemize}
        \item \textbf{Speedup Ratio ($SR$).} 
        The throughput ratio of speculative decoding to standard autoregressive generation, measured in tokens per second and indicating practical speedup.
        % The throughput ratio of speculative decoding to standard autoregressive generation, measured in tokens per second, reflecting the practical acceleration in deployment.
        \item \textbf{Average Acceptance Length ($\tau$).} The average number of drafter tokens accepted per verification step. Larger $\tau$ indicates stronger alignment with the target VLM and higher decoding efficiency.
    \end{itemize}
\textbf{Training Setup:}
    Consistent with the data configuration used in MSD \cite{lin2025speculative}, we mix 68,000 ShareGPT dialogues as the text-only dataset and 68,000 samples from LLaVA-Mix665k as the multimodal dataset. Only textual embeddings are retained, alleviating training overhead. Since many multimodal samples contain single-token answers, which is unsuitable for time-step aligned training, we filter out entries with answers shorter than five words.
    % Following MSD\cite{lin2025speculative}, for text-only training data, we use 68,000 dialogue instances from ShareGPT, while for multimodal dataset we sample 68,000 dialogues from the LLaVA-Mix665k dataset. 

    During training, we cap the draft length at $N=4$ to efficiently learn the bias-correction residuals $r^0$-$r^3$. At inference time, if drafting exceeds four steps, we reuse $r^3$ for all later steps. We use $K=5$ top candidates per step and set loss coefficients to $\beta=0.1$ and $\gamma=1$. All remaining hyperparameters follow EAGLE-2 \cite{li2024eagle2}.

\textbf{Inference Setup:}
% The experiments are conducted on a single A800 GPU with batch size set to 1. In the draft stage, the draft tree expands by selecting the top-10 candidates at each layer with a fixed depth $\lambda = 6$, and the top-60 paths ranked by cumulative probability product are retained as candidate outputs.
    Experiments are conducted on a single A800 GPU with batch size 1. During drafting, we expand a top-10 draft tree to depth $\lambda = 6$ and retain the top-60 paths ranked by cumulative probability.

\begin{figure*}[t]
    \centering
    \includegraphics[width=1\textwidth]{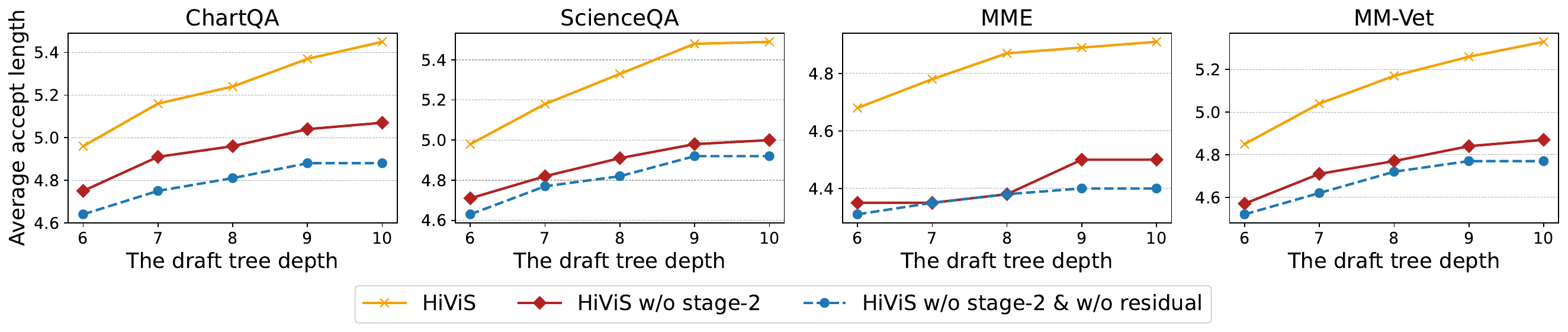}
    \vspace{-0.3cm}
    \caption{Ablation study of HiViS on four tasks evaluated on LLaVA-Next-7B under  different draft tree depths (6-10). The comparison includes: (1) full HiViS, (2) HiViS without stage-2 training, and (3) HiViS without both stage-2 and bias-correction residuals.}
    \label{fig:diff_depth_ablation}
    \vspace{-0.3cm}
\end{figure*}

\subsection{Results}
% \Cref{tab:depth5_result} reports a detailed comparison of the speedup ratio and average acceptance length across various VLMs and benchmarks. Despite removing all explicit visual information from both training and inference, HiViS consistently achieves higher speedup ratios than all baselines across models and tasks.
\Cref{tab:depth5_result} summarizes the speedup ratio and average acceptance length across multiple VLMs and tasks. Even without any explicit visual inputs during training or inference, HiViS consistently attains higher speedup ratios than all baselines.
At temperature $T=0$ (greedy sampling), HiViS achieves up to a $3.15\times$ speedup on Qwen2.5-VL-7B for TextVQA, exceeding ViSpec ($2.87\times$) and EAGLE-2 ($2.31\times$).
When sampling with $T=1$, HiViS maintains clear advantages, achieving speedups between $1.48\times$ and $2.39\times$, along with higher acceptance lengths, for example 4.38 on ChartQA with Qwen2.5-VL-7B, compared with 3.88 for ViSpec and 2.53 for EAGLE-2.
% Table~\ref{tab:depth3_result} further compares HiViS with the latest ViSpec framework~\cite{kang2025vispec} under its official configuration (identical system prompts and draft settings). HiViS remains fully compatible with this setup and continues to exhibit overall superiority, sustaining speedup ratios above $1.59\times$ at $T=1$ and reaching up to $3.04\times$ at $T=0$. 
Since ViSpec adopts additional system prompt, task-specific instructions, and different draft parameters, we further evaluate HiViS under their setting without additional training. Even so, HiViS still surpasses ViSpec across all tasks in both speedup ratio and average acceptance length. Full results are included in the Supplementary Material.

% These results collectively demonstrate that, in VLMs, the drafter for speculative decoding can operate effectively without any explicit visual-token input, and further validate the efficiency and robustness of the proposed training framework.
These results show that drafters for VLM can function effectively without explicit visual tokens, confirming the efficiency and robustness of our framework.

\subsection{Ablation Studies}
% We conduct ablation studies using LLaVA-Next-7B model with the temperature fixed at $0$. We analyze the effect of training stages and dataset composition on the drafter's performance, as measured by the average acceptance length $\tau$. Since our implementation does not include any additional latency-inducing modules, the drafter's capability can be regarded as approximately proportional to its average acceptance length.
We conduct studies on LLaVA-Next-7B at temperature $0$, examining how visual tokens, training stages and dataset composition affect the drafter, measured by average acceptance length $\tau$. 
%With no additional latency-inducing components, $\tau$ reliably reflects the relative drafter's performance.

% \begin{table}[h]
% \centering
% \begin{tabular}{|c|c|}
% \hline
% \textbf{training stage} & \textbf{$\tau$} \\ \hline
% stage-1 & 4.84 \\ \hline
% stage-2 & 5.20 \\ \hline
% \end{tabular}
% \caption{The average accept length after stage-1 training and stage-2 training}
% \label{abl:training stage}
% \end{table}

\textbf{Effect of visual tokens.}
% We compare two variants of the drafter with stage-1 training: one that receives the full sequence of visual and textual tokens, and another that takes only textual tokens as input, which is the HiViS design.
% We vary whether visual tokens and the training skill are included during stage-1 training. When trained only on the multimodal dataset, removing visual tokens already brings a clear improvement in both speedup ratio and acceptance length. After further adding the text-only dataset and training the bias-correction residual, performance improves substantially again. These results show that explicit visual tokens are unnecessary for effective speculative decoding.
We vary visual-token input and training configurations in stage-1 as shown in \Cref{tab:visual_input}.
Removing visual tokens already improves performance when trained on the multimodal dataset, and adding text-only data plus bias correction residual training yields a further boost. These results indicate that explicit visual tokens are unnecessary for effective speculative decoding.

% \begin{table}[h]
% \centering
% \small
% \begin{tabular}{ccccccc}
% \hline
% \multirow{2}{*}{\shortstack{stage-1 \\ input}} & \multicolumn{2}{c}{ChartQA} & \multicolumn{2}{c}{MM-Vet} & \multicolumn{2}{c}{ScienceQA} \\
% % \cline{2-7}
% & $SR$ & $\tau$ & $SR$ & $\tau$ & $SR$ & $\tau$ \\
% \hline
% \shortstack{visual \\ \& text} & & & & & & \\
% \hline
% text tokens & 4.75 & 1.82x & 4.57 & 1.87x & 4.71 & 2.01x\\
% \hline
% % \label{tab:visual_input}
% \end{tabular}
% \end{table}

\begin{table}[t]
\centering
\small
\caption{Stage-1 with different input sequences (“txt” for text, “vis” for visual) and training skills.}
\begin{tabular}{cccccccc}
\hline
\multirow{2}{*}{\shortstack{Input \\ tokens}} & \multirow{2}{*}{\shortstack{Training}} & \multicolumn{2}{c}{MME} & \multicolumn{2}{c}{MM-Vet} \\
% \cline{2-7}
&& $SR$ & $\tau$ & $SR$ & $\tau$ \\
\hline
\makecell{vis \& txt} & \makecell{multimodel data} & 1.33x & 3.35 & 1.42x & 2.81 \\
\hline
\makecell{txt} & \makecell{multimodel data} & 1.59x & 3.76 & 1.59x & 3.16 \\
\hline
\makecell{txt} & \makecell{+ text data \\ \& residual} & \textbf{1.82x} & \textbf{4.35} & \textbf{1.87x} & \textbf{4.57} \\
\hline
\label{tab:visual_input}
\end{tabular}
    \vspace{-0.3cm}
\end{table}

\textbf{Effect of time-step aligned training and bias correction residual.}
% \Cref{fig:diff_depth_ablation} shows the average acceptance length with draft tree depths ranging from 6 to 10. Across all tasks, HiViS without stage-2 training and bias-correction residual performs the worst. When the residual is added but not yet refined through stage-2, the performance notably improves, suggesting that the initial residual $r^0$ already contributes to stabilizing multi-step drafting. After stage-2 training, which further optimizes the residuals through time-step aligned supervision, HiViS achieves the highest and continuously rising acceptance length as the tree depth increases. This demonstrates that the proposed residual-related training strategy enables the drafter to maintain better alignment with the target VLM over longer drafting sequences. 
\Cref{fig:diff_depth_ablation} reports average acceptance length for draft tree depths 6 to 10. Across all tasks, HiViS without stage-2 training or residuals performs the worst. Adding residuals yields clear improvement, indicating that the initial $r^0$ already helps stabilize independent drafting. After stage-2 refinement, acceptance length becomes the highest and continues to grow with depth, showing that the residual-based, time-step aligned training strengthens drafter's alignment with the target VLM over longer drafting sequences.

\begin{table}[h]
\centering
\small
\caption{Effect of stage-1 training dataset composition on average acceptance length. Drafters trained with both multimodal and text-only data achieve the best performance across tasks.}
\begin{tabularx}{0.45\textwidth}{c|p{1.6cm}<{\centering}|p{1.6cm}<{\centering}}
\hline
Stage-1 Training Dataset & MME & MM-Vet \\ \hline
Multimodal only & 3.86 & 3.23 \\ \hline
Text only & 3.85 & 4.22 \\ \hline
Multimodal + text & \textbf{4.35} & \textbf{4.57} \\ \hline
\end{tabularx}
\label{tab:stage1_dataset}
    \vspace{-0.3cm}
\end{table}

\textbf{Effect of dataset composition.} We analyze the impact of stage-1 dataset composition across different tasks. As shown in \Cref{tab:stage1_dataset}, the performance varies when trained on multimodal or text-only data, reflecting their distinct contributions to different task types. Combining both datasets consistently achieves the best results, suggesting that the complementary strengths of multimodal and textual supervision enhance the drafter's robustness and generalization.

\section{Conclusion}
% We present \textbf{H}iding \textbf{V}isual Tokens from the Drafter for \textbf{S}peculative Decoding in Vision-Language Models (\textbf{HiViS}), a speculative decoding framework for VLMs that eliminates explicit visual token consumption while preserving multimodal grounding. By reformulating the drafter's conditioning into explicit textual inputs and implicit hidden-state guidance, DeViS avoids misaligned visual representations in the KV-cache and achieves a compact state representation that accelerates decoding under a lightweight architecture.
% To enhance alignment and robustness, we introduce a refined multi-step self-feedback training strategy that integrates dynamic data filtering and step-dependent sequential embeddings. This design enables efficient learning from within-capacity yet challenging examples, maintains stability over long drafting, and equips the drafter with the ability to autonomously propagate and update visual semantics during independent inference.  
% Comprehensive experiments on LLaVA-Next (7B and 13B) across multiple tasks demonstrate that HiViS consistently surpasses EAGLE-2 and MSD in both speedup ratio and acceptance length, despite retaining only $0.7\%-1.3\%$ of the input sequence length. These results establish speculative decoding without explicit visual tokens as a viable and effective paradigm for VLMs.
We present HiViS, a speculative decoding framework for VLMs that removes all visual tokens from the drafter while preserving multimodal semantics by using the target VLM as a natural semantic fusion module. HiViS further strengthens the drafter via time-step aligned training with step-dependent residuals. Across multiple VLMs and benchmarks, HiViS achieves the highest acceleration while maintaining strong alignment with the target VLM, demonstrating that speculative decoding for VLMs can be performed effectively without explicit visual tokens and paving the way for more lightweight drafter designs.

{
    \small
    \bibliographystyle{ieeenat_fullname}
    \bibliography{main}

@article{achiam2023gpt,
  title={Gpt-4 technical report},
  author={Achiam, Josh and Adler, Steven and Agarwal, Sandhini and Ahmad, Lama and Akkaya, Ilge and Aleman, Florencia Leoni and Almeida, Diogo and Altenschmidt, Janko and Altman, Sam and Anadkat, Shyamal and others},
  journal={arXiv preprint arXiv:2303.08774},
  year={2023}
}

@article{touvron2023llama,
  title={Llama: Open and efficient foundation language models},
  author={Touvron, Hugo and Lavril, Thibaut and Izacard, Gautier and Martinet, Xavier and Lachaux, Marie-Anne and Lacroix, Timoth{\'e}e and Rozi{\`e}re, Baptiste and Goyal, Naman and Hambro, Eric and Azhar, Faisal and others},
  journal={arXiv preprint arXiv:2302.13971},
  year={2023}
}

@inproceedings{leviathan2023fast,
  title={Fast inference from transformers via speculative decoding},
  author={Leviathan, Yaniv and Kalman, Matan and Matias, Yossi},
  booktitle={International Conference on Machine Learning},
  pages={19274--19286},
  year={2023},
  organization={PMLR}
}

@article{chen2023accelerating,
  title={Accelerating large language model decoding with speculative sampling},
  author={Chen, Charlie and Borgeaud, Sebastian and Irving, Geoffrey and Lespiau, Jean-Baptiste and Sifre, Laurent and Jumper, John},
  journal={arXiv preprint arXiv:2302.01318},
  year={2023}
}

@inproceedings{xia2022speculative,
    title = "Speculative Decoding: Exploiting Speculative Execution for Accelerating Seq2seq Generation",
    author = "Xia, Heming  and
      Ge, Tao  and
      Wang, Peiyi  and
      Chen, Si-Qing  and
      Wei, Furu  and
      Sui, Zhifang",
    editor = "Bouamor, Houda  and
      Pino, Juan  and
      Bali, Kalika",
    booktitle = "Findings of the Association for Computational Linguistics: EMNLP 2023",
    month = dec,
    year = "2023",
    address = "Singapore",
    publisher = "Association for Computational Linguistics",
    doi = "10.18653/v1/2023.findings-emnlp.257",
    pages = "3909--3925",
}

@article{fu2024break,
  title={Break the sequential dependency of llm inference using lookahead decoding},
  author={Fu, Yichao and Bailis, Peter and Stoica, Ion and Zhang, Hao},
  journal={arXiv preprint arXiv:2402.02057},
  year={2024}
}

@inproceedings{li2024eagle, 
	author = {Yuhui Li and Fangyun Wei and Chao Zhang and Hongyang Zhang}, 
	title = {{EAGLE}: Speculative Sampling Requires Rethinking Feature Uncertainty}, 
	booktitle = {International Conference on Machine Learning},
	year = {2024}
}

@inproceedings{li2024eagle2, 
	author = {Yuhui Li and Fangyun Wei and Chao Zhang and Hongyang Zhang}, 
	title = {{EAGLE-2}: Faster Inference of Language Models with Dynamic Draft Trees}, 
	booktitle = {Empirical Methods in Natural Language Processing},
	year = {2024}
}

@inproceedings{du2024glide,
author = {Du, Cunxiao and Jiang, Jing and Yuanchen, Xu and Wu, Jiawei and Yu, Sicheng and Li, Yongqi and Li, Shenggui and Xu, Kai and Nie, Liqiang and Tu, Zhaopeng and You, Yang},
title = {GLIDE with a CAPE: a low-hassle method to accelerate speculative decoding},
year = {2024},
publisher = {JMLR.org},
booktitle = {Proceedings of the 41st International Conference on Machine Learning},
articleno = {465},
numpages = {17},
location = {Vienna, Austria},
series = {ICML'24}
}

@article{sun2024triforce,
  title={Triforce: Lossless acceleration of long sequence generation with hierarchical speculative decoding},
  author={Sun, Hanshi and Chen, Zhuoming and Yang, Xinyu and Tian, Yuandong and Chen, Beidi},
  journal={arXiv preprint arXiv:2404.11912},
  year={2024}
}

@article{cai2024medusa,
  title={Medusa: Simple llm inference acceleration framework with multiple decoding heads},
  author={Cai, Tianle and Li, Yuhong and Geng, Zhengyang and Peng, Hongwu and Lee, Jason D and Chen, Deming and Dao, Tri},
  journal={arXiv preprint arXiv:2401.10774},
  year={2024}
}

@article{lin2025speculative,
  title={Speculative Decoding Reimagined for Multimodal Large Language Models},
  author={Lin, Luxi and Lin, Zhihang and Zeng, Zhanpeng and Ji, Rongrong},
  journal={arXiv preprint arXiv:2505.14260},
  year={2025}
}

@inproceedings{leebatch,
    title={In-batch Ensemble Drafting: Robust Speculative Decoding for {LVLM}s},
    author={Minjae Lee and Wonjun Kang and Byeongkeun Ahn and Christian Classen and Minghao Yan and Hyung Il Koo and Kangwook Lee},
    booktitle={First Workshop on Scalable Optimization for Efficient and Adaptive Foundation Models},
    year={2025}
}

@misc{vicuna2023,
    title = {Vicuna: An Open-Source Chatbot Impressing GPT-4 with 90\%* ChatGPT Quality},
    author = {Chiang, Wei-Lin and Li, Zhuohan and Lin, Zi and Sheng, Ying and Wu, Zhanghao and Zhang, Hao and Zheng, Lianmin and Zhuang, Siyuan and Zhuang, Yonghao and Gonzalez, Joseph E. and Stoica, Ion and Xing, Eric P.},
    month = {March},
    year = {2023}
}

@article{wang2024emu3,
  title={Emu3: Next-token prediction is all you need},
  author={Wang, Xinlong and Zhang, Xiaosong and Luo, Zhengxiong and Sun, Quan and Cui, Yufeng and Wang, Jinsheng and Zhang, Fan and Wang, Yueze and Li, Zhen and Yu, Qiying and others},
  journal={arXiv preprint arXiv:2409.18869},
  year={2024}
}

@article{wang2024qwen2,
  title={Qwen2-vl: Enhancing vision-language model's perception of the world at any resolution},
  author={Wang, Peng and Bai, Shuai and Tan, Sinan and Wang, Shijie and Fan, Zhihao and Bai, Jinze and Chen, Keqin and Liu, Xuejing and Wang, Jialin and Ge, Wenbin and others},
  journal={arXiv preprint arXiv:2409.12191},
  year={2024}
}

@article{team2024qwen2,
  title={Qwen2 technical report},
  author={Team, Qwen},
  journal={arXiv preprint arXiv:2407.10671},
  volume={2},
  year={2024}
}

@inproceedings{li2023blip,
  title={Blip-2: Bootstrapping language-image pre-training with frozen image encoders and large language models},
  author={Li, Junnan and Li, Dongxu and Savarese, Silvio and Hoi, Steven},
  booktitle={International conference on machine learning},
  pages={19730--19742},
  year={2023},
  organization={PMLR}
}

@misc{liu2023llava,
      title={Visual Instruction Tuning}, 
      author={Liu, Haotian and Li, Chunyuan and Wu, Qingyang and Lee, Yong Jae},
      publisher={NeurIPS},
      year={2023},
}

@misc{liu2024llavanext,
    title={LLaVA-NeXT: Improved reasoning, OCR, and world knowledge},
    author={Liu, Haotian and Li, Chunyuan and Li, Yuheng and Li, Bo and Zhang, Yuanhan and Shen, Sheng and Lee, Yong Jae},
    month={January},
    year={2024}
}

@article{zhang2024learning,
  title={Learning Harmonized Representations for Speculative Sampling},
  author={Zhang, Lefan and Wang, Xiaodan and Huang, Yanhua and Xu, Ruiwen},
  journal={arXiv preprint arXiv:2408.15766},
  year={2024}
}

@article{xu2024speculative,
  title={Speculative knowledge distillation: Bridging the teacher-student gap through interleaved sampling},
  author={Xu, Wenda and Han, Rujun and Wang, Zifeng and Le, Long T and Madeka, Dhruv and Li, Lei and Wang, William Yang and Agarwal, Rishabh and Lee, Chen-Yu and Pfister, Tomas},
  journal={arXiv preprint arXiv:2410.11325},
  year={2024}
}

@article{masry2022chartqa,
  title={Chartqa: A benchmark for question answering about charts with visual and logical reasoning},
  author={Masry, Ahmed and Long, Do Xuan and Tan, Jia Qing and Joty, Shafiq and Hoque, Enamul},
  journal={arXiv preprint arXiv:2203.10244},
  year={2022}
}

@article{lu2022learn,
  title={Learn to explain: Multimodal reasoning via thought chains for science question answering},
  author={Lu, Pan and Mishra, Swaroop and Xia, Tanglin and Qiu, Liang and Chang, Kai-Wei and Zhu, Song-Chun and Tafjord, Oyvind and Clark, Peter and Kalyan, Ashwin},
  journal={Advances in Neural Information Processing Systems},
  volume={35},
  pages={2507--2521},
  year={2022}
}

@inproceedings{singh2019towards,
  title={Towards vqa models that can read},
  author={Singh, Amanpreet and Natarajan, Vivek and Shah, Meet and Jiang, Yu and Chen, Xinlei and Batra, Dhruv and Parikh, Devi and Rohrbach, Marcus},
  booktitle={Proceedings of the IEEE/CVF conference on computer vision and pattern recognition},
  pages={8317--8326},
  year={2019}
}

@inproceedings{goyal2017making,
  title={Making the v in vqa matter: Elevating the role of image understanding in visual question answering},
  author={Goyal, Yash and Khot, Tejas and Summers-Stay, Douglas and Batra, Dhruv and Parikh, Devi},
  booktitle={Proceedings of the IEEE conference on computer vision and pattern recognition},
  pages={6904--6913},
  year={2017}
}

@article{zhang2024cls,
  title={[CLS] Attention is All You Need for Training-Free Visual Token Pruning: Make VLM Inference Faster},
  author={Zhang, Qizhe and Cheng, Aosong and Lu, Ming and Zhuo, Zhiyong and Wang, Minqi and Cao, Jiajun and Guo, Shaobo and She, Qi and Zhang, Shanghang},
  journal={arXiv e-prints},
  pages={arXiv--2412},
  year={2024}
}

@article{zhang2025vispruner,
      title={Beyond Text-Visual Attention: Exploiting Visual Cues for Effective Token Pruning in VLMs}, 
      author={Zhang, Qizhe and Cheng, Aosong and Lu, Ming and Zhang, Renrui and Zhuo, Zhiyong and Cao, Jiajun and Guo, Shaobo and She, Qi and Zhang, Shanghang},
      journal={arXiv preprint arXiv:2412.01818},
      year={2025},
}

@article{bai2025qwen2,
  title={Qwen2. 5-vl technical report},
  author={Bai, Shuai and Chen, Keqin and Liu, Xuejing and Wang, Jialin and Ge, Wenbin and Song, Sibo and Dang, Kai and Wang, Peng and Wang, Shijie and Tang, Jun and others},
  journal={arXiv preprint arXiv:2502.13923},
  year={2025}
}

@misc{fu2025mme,
      title={MME: A Comprehensive Evaluation Benchmark for Multimodal Large Language Models}, 
      author={Chaoyou Fu and Peixian Chen and Yunhang Shen and Yulei Qin and Mengdan Zhang and Xu Lin and Jinrui Yang and Xiawu Zheng and Ke Li and Xing Sun and Yunsheng Wu and Rongrong Ji and Caifeng Shan and Ran He},
      year={2025},
      eprint={2306.13394},
      archivePrefix={arXiv},
      primaryClass={cs.CV}
}

@article{yu2023mm,
  title={Mm-vet: Evaluating large multimodal models for integrated capabilities},
  author={Yu, Weihao and Yang, Zhengyuan and Li, Linjie and Wang, Jianfeng and Lin, Kevin and Liu, Zicheng and Wang, Xinchao and Wang, Lijuan},
  journal={arXiv preprint arXiv:2308.02490},
  year={2023}
}

@inproceedings{kang2025vispec,
  title={ViSpec: Accelerating Vision-Language Models with Vision-Aware Speculative Decoding},
  author={Kang, Jialiang and Shu, Han and Li, Wenshuo and Zhai, Yingjie and Chen, Xinghao},
  booktitle={Annual Conference on Neural Information Processing Systems},
  year={2025}
}

@article{tu2024vl,
  title={VL-cache: Sparsity and modality-aware KV cache compression for vision-language model inference acceleration},
  author={Tu, Dezhan and Vashchilenko, Danylo and Lu, Yuzhe and Xu, Panpan},
  journal={arXiv preprint arXiv:2410.23317},
  year={2024}
}

@article{huang2024dynamic,
  title={Dynamic-llava: Efficient multimodal large language models via dynamic vision-language context sparsification},
  author={Huang, Wenxuan and Zhai, Zijie and Shen, Yunhang and Cao, Shaosheng and Zhao, Fei and Xu, Xiangfeng and Ye, Zheyu and Hu, Yao and Lin, Shaohui},
  journal={arXiv preprint arXiv:2412.00876},
  year={2024}
}

@inproceedings{
zhang2025llava,
title={{LL}a{VA}-Mini: Efficient Image and Video Large Multimodal Models with One Vision Token},
author={Shaolei Zhang and Qingkai Fang and Zhe Yang and Yang Feng},
booktitle={The Thirteenth International Conference on Learning Representations},
year={2025}
}

@inproceedings{chen2024image,
  title={An image is worth 1/2 tokens after layer 2: Plug-and-play inference acceleration for large vision-language models},
  author={Chen, Liang and Zhao, Haozhe and Liu, Tianyu and Bai, Shuai and Lin, Junyang and Zhou, Chang and Chang, Baobao},
  booktitle={European Conference on Computer Vision},
  pages={19--35},
  year={2024},
  organization={Springer}
}

@article{li2023seed,
  title={Seed-bench: Benchmarking multimodal llms with generative comprehension},
  author={Li, Bohao and Wang, Rui and Wang, Guangzhi and Ge, Yuying and Ge, Yixiao and Shan, Ying},
  journal={arXiv preprint arXiv:2307.16125},
  year={2023}
}

@inproceedings{hudson2019gqa,
  title={Gqa: A new dataset for real-world visual reasoning and compositional question answering},
  author={Hudson, Drew A and Manning, Christopher D},
  booktitle={Proceedings of the IEEE/CVF conference on computer vision and pattern recognition},
  pages={6700--6709},
  year={2019}
}

@article{hu2025dream,
  title={DREAM: Drafting with Refined Target Features and Entropy-Adaptive Cross-Attention Fusion for Multimodal Speculative Decoding},
  author={Hu, Yunhai and Xia, Tianhua and Liu, Zining and Raman, Rahul and Liu, Xingyu and Bao, Bo and Sather, Eric and Thangarasa, Vithursan and Zhang, Sai Qian},
  journal={arXiv preprint arXiv:2505.19201},
  year={2025}
}
}

\clearpage
% \setcounter{page}{1}
% \maketitlesupplementary

\onecolumn

\section{Evaluation Under ViSpec Settings}

we additionally evaluate HiViS and our reproduced EAGLE-2 under the exact settings used by ViSpec. Following their protocol, we prepend the same system prompt: \emph{A chat between a curious human and an artificial intelligence assistant. The assistant gives helpful, detailed, and polite answers to the human's questions.} and include all task-specific instructions provided in the dataset. During speculative decoding, we adopt ViSpec's draft token tree configuration, using a total of 30 draft tokens, a tree depth of 4, and 8 nodes selected during each expansion step. Under this evaluation setup, without any additional training, HiViS still consistently surpasses ViSpec in both speedup ratio and average acceptance length across all tested tasks. Full results are reported in \Cref{tab:depth3_result}.

\begin{table*}[h]
\centering
\scriptsize
\caption{Speedup ratio ($SR$) and average acceptance length ($\tau$) across benchmarks under ViSpec's settings.}
\setlength{\tabcolsep}{2pt}
\begin{tabular}{@{}l l cc cc cc cc cc cc cc cc cc@{}}
\toprule
\multirow{2}{*}{Model} & \multirow{2}{*}{Methods} & \multicolumn{2}{c}{ChartQA}  & \multicolumn{2}{c}{VQAv2}  & \multicolumn{2}{c}{ScienceQA}  & \multicolumn{2}{c}{TextVQA}  & \multicolumn{2}{c}{MME}  & \multicolumn{2}{c}{MMVet} & \multicolumn{2}{c}{SEED-Bench} & \multicolumn{2}{c}{GQA} & \multicolumn{2}{c}{Avg} \\
&  & $SR$ & $\tau$  & $SR$ & $\tau$  & $SR$ & $\tau$  & $SR$ & $\tau$  & $SR$ & $\tau$  & $SR$ & $\tau$  & $SR$ & $\tau$ & $SR$ & $\tau$ & $SR$ & $\tau$ \\
\midrule
\multicolumn{20}{c}{T = 0} \\
\midrule
\multirow{3}{*}{LLaVA-Next-7B} & EAGLE-2 & 1.37$\times$ & 2.75 & 1.94$\times$ & 3.66 & 1.52$\times$ & 2.80 & 1.52$\times$ & 2.96 & 1.53$\times$ & 2.91 & 1.50$\times$ & 2.80 & 1.67$\times$  & 3.64  & 1.55$\times$ & 3.56 & 1.58$\times$  & 3.14  \\
 & ViSpec & 1.75$\times$ & 3.98 & 2.02$\times$ & 3.82 & 2.36$\times$ & 3.87 & 2.10$\times$ & 3.99 & 1.84$\times$ & 3.80 & 1.81$\times$ & 3.81  & 1.78$\times$  & 4.08  & 1.63$\times$  & 3.86  & 1.91$\times$  & 3.90 \\
 & HiViS & \textbf{1.80$\times$} & \textbf{4.03} & \textbf{2.19$\times$} & \textbf{4.34} & \textbf{2.52$\times$} & \textbf{4.05} & \textbf{2.19$\times$} & \textbf{4.07} & \textbf{1.89$\times$} & \textbf{3.81} & \textbf{1.90$\times$} & \textbf{3.96} & \textbf{1.85$\times$} & \textbf{4.17} & \textbf{1.72$\times$} & \textbf{4.15} & \textbf{2.01$\times$} & \textbf{4.07} \\
\midrule
\multirow{3}{*}{LLaVA-Next-13B} & EAGLE-2 & 1.50$\times$ & 2.78 & 1.80$\times$ & 3.67 & 1.69$\times$ & 2.80 & 1.63$\times$ & 2.08 & 1.66$\times$ & 2.92 & 1.63$\times$ & 2.82 & 1.75$\times$  & 3.47  & 1.72$\times$  & 3.61  & 1.67$\times$  & 3.02 \\
 & ViSpec & 1.84$\times$ & 3.91 & 2.47$\times$ & 3.88 & 2.39$\times$ & 3.71 & 2.16$\times$ & 3.78 & 2.00$\times$ & 3.83 & 1.93$\times$ & 3.66 & 1.94$\times$  & 4.08  & 1.83$\times$  & 3.93  & 2.07$\times$  & 3.85 \\
 & HiViS & \textbf{1.88$\times$} & \textbf{4.01} & \textbf{2.61$\times$} & \textbf{4.27} & \textbf{2.56$\times$} & \textbf{4.00} & \textbf{2.32$\times$} & \textbf{4.03} & \textbf{2.07$\times$} & \textbf{3.93} & \textbf{2.02$\times$} & \textbf{3.91} & \textbf{1.97$\times$} & \textbf{4.13} & \textbf{1.87$\times$} & \textbf{4.22} & \textbf{2.16$\times$} & \textbf{4.06} \\
\midrule
\multirow{3}{*}{Qwen2.5-VL-7B} & EAGLE-2 & 1.46$\times$  & 2.44  & 2.00$\times$  & 3.47  & 1.57$\times$  & 2.57  & 2.34$\times$  & 2.44  & 1.58$\times$  & 2.54  & 1.53$\times$  & 2.37  & 1.71$\times$  & 3.06  & 1.99$\times$  & 3.43  & 1.77$\times$  & 2.79 \\
 & ViSpec & 1.85$\times$ & 3.56 & 1.88$\times$ & 3.22 & 1.87$\times$ & 3.17 & 2.94$\times$ & 3.22 & 1.80$\times$ & 3.09 & 1.78$\times$ & 3.08  & 1.74$\times$  & 3.25  & 1.93$\times$  & 3.30  & 1.97$\times$  & 3.24 \\
 % & HiViS (qwen) & 1.98$\times$ & 3.81 & 2.15$\times$ & 3.76 & 2.03$\times$ & 3.41 & 3.01$\times$ & 3.35 & 1.90$\times$ & 3.23 & 1.91$\times$ & 3.30 & 2.16$\times$ & 3.48 \\
 & HiViS & \textbf{1.97$\times$} & \textbf{3.85} & \textbf{2.13$\times$} & \textbf{3.78} & \textbf{2.04$\times$} & \textbf{3.47} & \textbf{3.04$\times$} & \textbf{3.34} & \textbf{1.89$\times$} & \textbf{3.25} & \textbf{1.89$\times$} & \textbf{3.30} & \textbf{1.85$\times$}  & \textbf{3.51}  & \textbf{2.15$\times$}  & \textbf{3.74}  & \textbf{2.12$\times$}  & \textbf{3.53} \\
\midrule
\multicolumn{20}{c}{T = 1} \\
\midrule
\multirow{3}{*}{LLaVA-Next-7B} & EAGLE-2 & 1.22$\times$ & 2.34 & 1.48$\times$ & 2.92 & 1.30$\times$ & 2.33 & 1.21$\times$ & 2.50 & 1.31$\times$ & 2.38 & 1.30$\times$ & 2.46 & 1.44$\times$  & 2.80  & 1.42$\times$  & 2.79  & 1.34$\times$  & 2.57 \\
 & ViSpec & 1.45$\times$ & 3.18 & 1.82$\times$ & 3.07 & 1.93$\times$ & 2.96 & 1.50$\times$ & 2.84 & 1.57$\times$ & 2.99 & 1.51$\times$ & 3.01 & 1.53$\times$  & 3.15  & 1.43$\times$  & 3.13  & 1.59$\times$  & 3.04 \\
 & HiViS & \textbf{1.60$\times$} & \textbf{3.30} & \textbf{1.88$\times$} & \textbf{3.26} & \textbf{2.14$\times$} & \textbf{3.06} & \textbf{1.62$\times$} & \textbf{2.84} & \textbf{1.63$\times$} & \textbf{3.04} & \textbf{1.60$\times$} & \textbf{3.06} & \textbf{1.55$\times$} & \textbf{3.16} & \textbf{1.50$\times$} & \textbf{3.26} & \textbf{1.69$\times$} & \textbf{3.12} \\
\midrule
\multirow{3}{*}{LLaVA-Next-13B} & EAGLE-2 & 1.42$\times$ & 2.50 & 1.60$\times$ & 3.01 & 1.56$\times$ & 2.29 & 1.42$\times$ & 2.62 & 1.51$\times$ & 2.50 & 1.50$\times$ & 2.47 & 1.58$\times$  & 2.85  & 1.47$\times$  & 2.96  & 1.51$\times$  & 2.65 \\
 & ViSpec & 1.69$\times$ & 3.28 & 2.14$\times$ & 3.19 & 1.85$\times$ & 2.94 & 1.73$\times$ & 2.97 & 1.73$\times$ & 3.09 & 1.77$\times$ & 3.07 & 1.69$\times$  & 3.23  & 1.64$\times$  & 3.17  & 1.78$\times$  & 3.12 \\
  & HiViS & \textbf{1.76$\times$} & \textbf{3.39} & \textbf{2.28$\times$} & \textbf{3.40} & \textbf{2.16$\times$} & \textbf{3.13} & \textbf{1.83$\times$} & \textbf{3.07} & \textbf{1.78$\times$} & \textbf{3.16} & \textbf{1.81$\times$} & \textbf{3.20} & \textbf{1.89$\times$} & \textbf{3.29} & \textbf{1.75$\times$} & \textbf{3.37} & \textbf{1.91$\times$} & \textbf{3.25} \\
\midrule
\multirow{3}{*}{Qwen2.5-VL-7B} & EAGLE-2 & 1.43$\times$  & 2.42  & 1.63$\times$  & 2.64  & 1.39$\times$  & 2.28  & 1.79$\times$  & 1.99  & 1.35$\times$  & 2.14  & 1.33$\times$  & 2.04  & 1.40$\times$  & 2.41  & 1.61$\times$  & 2.68  & 1.49$\times$  & 2.33 \\
 & ViSpec & 1.78$\times$ & 3.41 & 1.57$\times$ & 2.65 & 1.57$\times$ & 2.67 & 2.15$\times$ & 2.44 & 1.51$\times$ & 2.47 & 1.54$\times$ & 2.50 & 1.39$\times$  & 2.54  & 1.57$\times$  & 2.63  & 1.64$\times$  & 2.66 \\
 % & HiViS (qwen) & 1.93$\times$ & 3.60 & 1.73$\times$ & 2.83 & 1.74$\times$ & 2.87 & 2.21$\times$ & 2.49 & 1.60$\times$ & 2.63 & 1.65$\times$ & 2.64 & 1.81$\times$ & 2.84 \\
 & HiViS & \textbf{1.89$\times$} & \textbf{3.61} & \textbf{1.74$\times$} & \textbf{2.92} & \textbf{1.73$\times$} & \textbf{2.88} & \textbf{2.19$\times$} & \textbf{2.47} & \textbf{1.59$\times$} & \textbf{2.59} & \textbf{1.63$\times$} & \textbf{2.65} & \textbf{1.43$\times$} & \textbf{2.74} & \textbf{1.72$\times$} & \textbf{2.93} & \textbf{1.74$\times$} & \textbf{2.85} \\
\midrule
\bottomrule
\end{tabular}
\label{tab:depth3_result}
\end{table*}

% \clearpage

\section{Mixed vs. All-Multimodal Training for HiViS}

\begin{figure*}[h]
    \centering
    \includegraphics[width=0.88\textwidth]{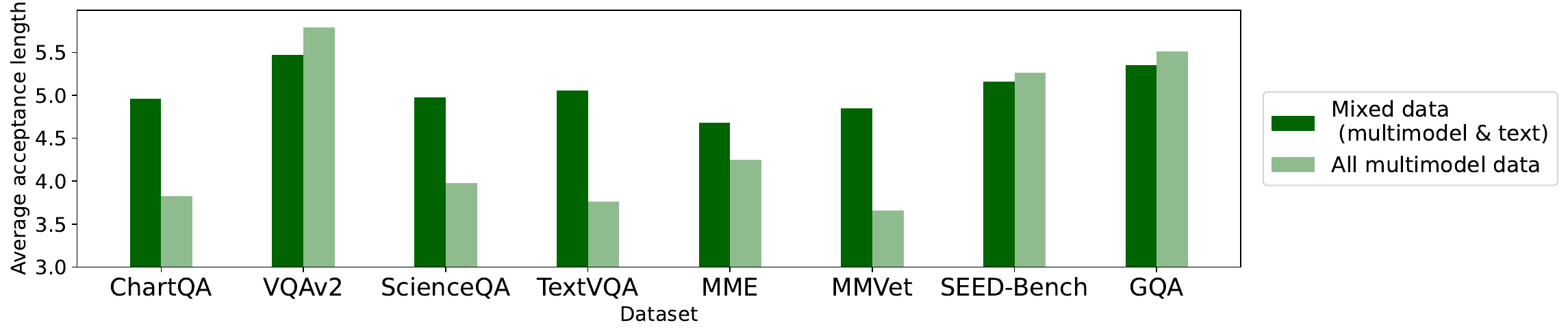}
    \caption{Comparison of average acceptance length between HiViS trained on mixed data (multimodal + text) and HiViS trained on multimodal-only data across several benchmarks, evaluated with a draft tree depth of 6 and temperature of 0.}
    \label{fig:dataset_compare}
\end{figure*}

We conduct an additional experiment by replacing the text-only dataset in HiViS's training with an equally size of samples' multimodal dataset. As shown in \Cref{fig:dataset_compare}, the mixed-dataset HiViS (multimodal + text) outperforms the all-multimodal variant on most tasks, and maintains only small gaps on the remaining ones. This outcome reflects a property of HiViS: since the drafter in HiViS operates purely in the fused language space and never observes raw visual tokens, its performance is primarily governed by how well it models the target VLM's next-token distribution rather than how well it processes visual sequences. Text-only dataset, which contain longer sequences and a much richer vocabulary, provides stronger supervision for learning long-range language dependencies, which in turn yields a drafter that is more robust across tasks.

% \section{Rationale}
% \label{sec:rationale}
% % 
% Having the supplementary compiled together with the main paper means that:
% % 
% \begin{itemize}
% \item The supplementary can back-reference sections of the main paper, for example, we can refer to \cref{sec:intro};
% \item The main paper can forward reference sub-sections within the supplementary explicitly (e.g. referring to a particular experiment); 
% \item When submitted to arXiv, the supplementary will already included at the end of the paper.
% \end{itemize}
% % 
% To split the supplementary pages from the main paper, you can use \href{https://support.apple.com/en-ca/guide/preview/prvw11793/mac#:~:text=Delete%20a%20page%20from%20a,or%20choose%20Edit%20%3E%20Delete).}{Preview (on macOS)}, \href{https://www.adobe.com/acrobat/how-to/delete-pages-from-pdf.html#:~:text=Choose%20%E2%80%9CTools%E2%80%9D%20%3E%20%E2%80%9COrganize,or%20pages%20from%20the%20file.}{Adobe Acrobat} (on all OSs), as well as \href{https://superuser.com/questions/517986/is-it-possible-to-delete-some-pages-of-a-pdf-document}{command line tools}.

\end{document}